\documentclass[sigconf]{acmart}

\usepackage{grffile}
\usepackage{amsmath}
\usepackage{tikz}
\usetikzlibrary{shapes,decorations,shapes.geometric,calc,arrows,arrows.meta,matrix,shapes.arrows,patterns}
\usepackage{siunitx}
\usepackage{threeparttable}
\usepackage{subcaption}
\usepackage{array}
\usepackage{mathrsfs}
\usepackage{textgreek}
\usepackage{pgfplots}
\usepackage{xstring}
\usepackage{enumitem}
\usepackage{multirow}
\usepackage{makecell}
\usepackage{color,soul}
\usepackage{amsthm}
\usepackage[T1]{fontenc}
\usepackage{algorithm2e}
\usepackage{algorithmic}

\usepackage{microtype}
\usepackage{graphicx}
\usepackage{booktabs} 

\usepackage{hyperref}

\newcommand\sys{AdaPM}
\newcommand\figurespace{\vspace{-6pt}}

\newcommand\rectM{$10\text{m}\times1\text{m}$}

\newcommand\add[1]{#1}

\newcommand\panelfiguresetup{
  \captionsetup[subfigure]{aboveskip=3pt,belowskip=6pt}
}

\definecolor{Worker1}{HTML}{1f78b4}
\definecolor{Worker1light}{HTML}{a6cee3}
\definecolor{Worker2}{HTML}{33a02c}
\definecolor{Worker2light}{HTML}{b2df8a}
\definecolor{Worker3}{HTML}{DCE319}
\definecolor{neutralbg}{HTML}{eeeeee}

\definecolor{Key3}{HTML}{481769}
\definecolor{Key4}{HTML}{D8E219}

\definecolor{Vir1}{HTML}{440154}
\definecolor{Vir2}{HTML}{481568}
\definecolor{Vir3}{HTML}{482677}
\definecolor{Vir4}{HTML}{453781}
\definecolor{Vir5}{HTML}{3F4788}
\definecolor{Vir6}{HTML}{39558C}
\definecolor{Vir7}{HTML}{32648E}
\definecolor{Vir8}{HTML}{2D718E}
\definecolor{Vir9}{HTML}{287D8E}
\definecolor{Vir10}{HTML}{238A8D}
\definecolor{Vir11}{HTML}{1F968B}
\definecolor{Vir12}{HTML}{20A386}
\definecolor{Vir13}{HTML}{29AF7F}
\definecolor{Vir14}{HTML}{3CBC75}
\definecolor{Vir15}{HTML}{56C667}
\definecolor{Vir16}{HTML}{74D055}
\definecolor{Vir17}{HTML}{94D840}
\definecolor{Vir18}{HTML}{B8DE29}
\definecolor{Vir19}{HTML}{DCE318}
\definecolor{Vir20}{HTML}{FDE725}

\def \routfontsize {\normalsize}

\tikzset{
	square/.style={regular polygon,regular polygon sides=4,inner sep=0mm}, 
	object/.style={square,text width = 0.9cm,draw=black, align=center,fill=neutralbg, rounded corners,font=\routfontsize}, 
	object_outlier/.style={square,text width = 0.9cm,draw=white, align=center,fill=white, text=white,rounded corners,font=\routfontsize}, 
	arrow_style/.style={->,>=stealth',line width=0.2mm, shorten <=4pt, shorten >=4pt}, 
	arrow_caption/.style={color=black,align=center,font=\routfontsize },
	description/.style={align=left, font=\routfontsize}
}

\newcommand{\policyplotbase}[1]{
  \begin{tikzpicture}
    [
    line width = 0.2mm,
    scale=0.30,
    alloc/.style={fill=#1!80,draw=none}, 
    replica/.style={fill=Worker1!50,draw=none}, 
    intent/.style={pattern=crosshatch,draw=none}, 
    replica2/.style={pattern=north east lines,draw=#1,pattern color=#1!80}, 
    ]
}

\newcommand{\policyplotstart}[3]{
  \policyplotbase{#1}

    \draw [->,>=stealth',line width=0.3mm,draw=black!50] (0,-0.5)--(#2,-0.5);
    \node [anchor=north east] at (#2-0.6,-0.5) {\scriptsize time};

    \pgfmathtruncatemacro{\yto}{#3 - 1}

    \foreach \y in {0,...,\yto}
    \node [anchor=east] at (0, \y+0.5) {\scriptsize node \y};

    \foreach \y in {0,...,\yto}
    \draw[thin, gray!20] (0,\y+0.5) -- ++(#2,0);

}

\newcommand{\policyplotend}[0]{
  \end{tikzpicture}
}

\newcommand{\pspolicy}[1]{
  \policyplotstart{Worker2}{10}{3}
}

\newcommand{\adapolicy}[1]{
  \policyplotstart{#1}{12}{4}
}

\newcommand{\adapolicyhalf}[1]{
  \policyplotstart{#1}{10}{4}
}

\def\accfour{5.6}

\newcommand{\alloc}[3]{
  \draw[alloc] ([yshift=0.5mm]#2,#1) rectangle ++([yshift=-2*0.5mm]#3,1);
}
\newcommand{\rep}[3]{
  \draw[replica] ([yshift=0.5mm]#2,#1) rectangle ++([yshift=-2*0.5mm]#3,1);
}
\newcommand{\preprep}[3]{

  \rep{#1}{#2-0.5}{#3+0.5}
}
\newcommand{\intent}[3]{

  \draw[thick] (#2,#1+0.5) -- ++(#3,0);

}

\newcommand{\intentsignal}[3]{
  \draw[very thick, gray] (#2,#1+0.5+0.25) -- ++(0,-0.5);
}

\newcommand{\actionwindow}[3]{
  \draw[thick, gray, dotted] (#2+0.2,#1+0.5) -- ++(#3-0.2,0);
}

\newcommand{\signal}[3]{
  \intentsignal{#1}{#2}{#3}
  \actionwindow{#1}{#2}{#3}
}

\newcommand{\sigintent}[3]{
  \customsigintent{#1}{#2}{#3}{1.5}
}

\newcommand{\customsigintent}[4]{
  \signal{#1}{#2-#4}{#4}
  \intent{#1}{#2}{#3}
}

\newcommand{\comm}[4]{
  \foreach \i in {#3,...,#4}{
    \draw[thick,red,{Stealth[length=0.7mm,width=1mm]}-{Stealth[length=0.7mm,width=1mm]}] (\i,#1+0.7) -- (\i,#2+0.3);
  }
}

\def\respstep{5}

\newcommand{\responsibility}[3]{
	\pgfmathsetmacro{\istart}{#2*\respstep}
	\pgfmathsetmacro{\iend}{(#2+#3)*\respstep}
  \foreach \i in {\istart,...,\iend}{
    \draw[thick,mypurple] (\i/\respstep,#1+0.5+0.18) -- ++(0,-0.36);
  }
}

\newcommand{\lacc}[2]{
  \pacc{#1}{#2}{black}
}

\newcommand{\pacc}[3]{
  \draw[thick,draw=#3,line width=0.4mm] ([yshift=2mm]#2,#1) -- ++([yshift=-4mm]0,1);
}

\newcommand{\lacchome}[1]{
  \lacc{0}{\accfour}
}

\tikzset{	
	col_object/.style={minimum width=4.9cm, minimum height = 2.8cm,draw=black, rounded corners, fill=neutralbg, align=center}, 
	col_arrow_style/.style={<->,>=stealth',line width=0.2mm, color=black, shorten <=4pt, shorten >=4pt}, 
	col_arrow_caption/.style={sloped, color=black,align=center,font=\scriptsize }, 
	col_rectangles/.style={draw=black,fill=Worker2light, rounded corners=0.8mm,minimum height=4.5mm,minimum width=3.2cm]},
	col_small_arrows/.style={<->,>=stealth', line width=0.26mm, shorten <=2pt, shorten >=2pt}
}

\def \colfontsize {\large}

\def \iall {17}
\def \ihalf {8}
\def \ihalfpp {9}

\def \yoffset {6.5}
\def \xoffset {4.5}

\def \textsep {1.1}

\newcommand{\drawP}[5]{
  \pgfmathsetmacro\y{int(#1/3)+#2*\yoffset}
  \pgfmathsetmacro\x{int(mod(#1,3))+#3*\xoffset}
  \draw [fill=#4, draw=#5] (\x,\y) rectangle (\x+1,\y+1);
}

\newcommand{\allocP}[3]{
  \drawP{#1}{#2}{#3}{Worker2}{black}
}

\newcommand{\repPsingle}[3]{
  \drawP{#1}{#2}{#3}{Worker1!50}{black}
}

\newcommand{\allocPstatic}[2]{
  \allocP{#1}{#2}{0}
  \allocP{#1}{#2}{1}
  \allocP{#1}{#2}{2}
}

\newcommand{\allocPall}[2]{
  \allocP{#1}{#2}{0}
  \allocP{#1}{#2}{1}
  \allocP{#1}{#2}{2}
}

\newcommand{\repPstatic}[2]{
  \repPsingle{#1}{#2}{0}
  \repPsingle{#1}{#2}{1}
  \repPsingle{#1}{#2}{2}
}

\newcommand{\baseP}[3]{
  \drawP{#1}{#2}{#3}{gray!15}{gray!35}
}

\newcommand{\approachStart}[2]{
  \begin{tikzpicture}[auto, scale=0.178,
    label/.style={align=center, anchor=center, minimum width=3, minimum height=1cm, color=gray},
    label/.append style={font=\vphantom{Ag}\scriptsize}
    ]
    \foreach \i in {0,...,\iall}
    \baseP{\i}{0}{0}
    \baseP{\i}{0}{1}
    \baseP{\i}{0}{2}
    ;
    \foreach \i in {0,...,\iall}
    \baseP{\i}{1}{0}
    \baseP{\i}{1}{1}
    \baseP{\i}{1}{2}
    ;

    \draw [->, >=stealth',auto,anchor=center] (3.2,6.25) to ++(1,0);
    \draw [->, >=stealth',auto] (7.7,6.25) to ++(1,0);

    \node [label] at (1.5, -\textsep) {initial};
    \node [label] at (\xoffset+1.5, -\textsep) {early};
    \node [label] at (2*\xoffset+1.5, -\textsep) {late};

    \node [label, rotate=90] at (-\textsep, 2.5) {node 1};
    \node [label, rotate=90] at (-\textsep, \yoffset+2.5) {node 2};
}

\newcommand{\approachEnd}[2]{

  \foreach \i in {0,...,2} {
    \draw [thick] (\i*\xoffset,0) rectangle ++(3,6);
    \draw [thick] (\i*\xoffset,\yoffset) rectangle ++(3,6);
  }

\end{tikzpicture}
}

\definecolor{primary}{HTML}{33a02c}
\definecolor{secondary}{HTML}{1f78b4}
\definecolor{third}{HTML}{ff7f00}
\definecolor{bg}{HTML}{999999}
\definecolor{darkred}{HTML}{DC143C}
\definecolor{mypurple}{HTML}{6a3d9a}

\newif\iflong
\longtrue 

\AtBeginDocument{%
	}

\acmBooktitle{To be published in CIKM '23}

\acmYear{2023}

\begin{document}

\title[Good Intentions: Adaptive Parameter Management via Intent Signaling]{Good Intentions: \\ Adaptive Parameter Management via Intent Signaling}

\author{Alexander Renz-Wieland}
\affiliation{%
  \institution{Technische Universität Berlin}
}

\author{Andreas Kieslinger}
\affiliation{%
  \institution{Technische Universität Berlin}
}
\affiliation{%
  \institution{BIFOLD}
}

\author{Robert Gericke}
\affiliation{%
  \institution{Technische Universität Berlin}
}
\affiliation{%
  \institution{BIFOLD}
}

\author{Rainer Gemulla}
\affiliation{%
  \institution{Universität Mannheim}
}

\author{Zoi Kaoudi}
\affiliation{%
  \institution{IT University Copenhagen}
}

\author{Volker Markl}
\affiliation{%
  \institution{Technische Universität Berlin}
}
\affiliation{%
  \institution{BIFOLD}
}

\begin{abstract}
  Model parameter management is essential for distributed training of large
  machine learning (ML) tasks. Some ML tasks are hard to distribute because
  common approaches to parameter management can be highly inefficient. Advanced
  parameter management approaches---such as selective replication or dynamic
  parameter allocation---can improve efficiency, but they typically need to be
  integrated manually into each task's implementation and they require expensive
  upfront experimentation to tune correctly. In this work, we explore whether
  these two problems can be avoided. We first propose a novel intent signaling
  mechanism that integrates naturally into existing ML stacks and provides the
  parameter manager with crucial information about parameter accesses. We then
  describe \sys{}, a fully adaptive, zero-tuning parameter manager based on this
  mechanism. \add{In contrast to prior parameter managers, our approach
    decouples how access information is provided (simple) from how and when it
    is exploited (hard). } In our experimental evaluation, \sys{} matched or
  outperformed state-of-the-art parameter managers out of the box, suggesting
  that automatic parameter management is possible.
\end{abstract}



\maketitle

\section{Introduction}
\label{sec:intro}

Distributed training is essential for large-scale machine learning (ML) tasks.
It enables (i) training of models that exceed the memory capacity of a single
node and (ii) faster training by leveraging the compute of multiple nodes of a
cluster. Typically, each node accesses one (local) partition of the training
data, but requires global read and write access to all model parameters. Thus,
\emph{parameter management} (PM) among nodes is a key concern in distributed
training. We refer to any system that provides distributed PM, i.e., that
provides global parameter access across a cluster, as \emph{parameter manager}.
Most ML systems include a parameter manager as core
component~\cite{tensorflow, pytorch, mxnet}.

\begin{figure}
  \centering
  \includegraphics[page=1,width=1\columnwidth]{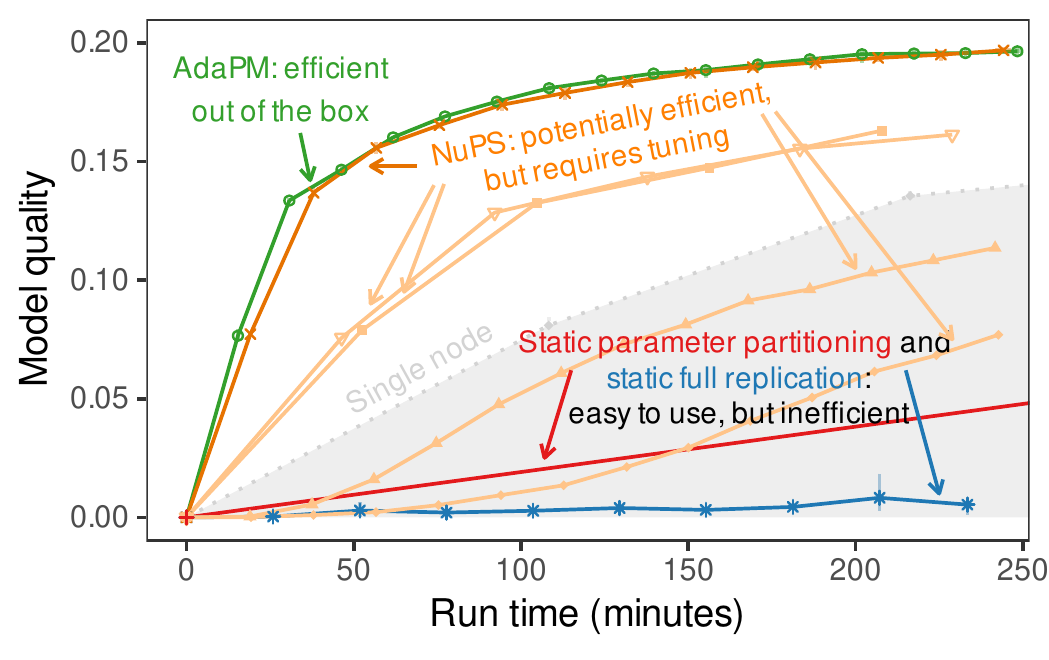}
  \caption{Performance for training large knowledge graph embeddings
    (ComplEx, dimension~500 on Wikidata5m) on an 8-node
    cluster. Static full replication and static parameter partitioning are easy
    to use, but inefficient. NuPS can be more efficient, but is hard to use.
    \sys{} is easy to use and efficient. Details in
    Section~\ref{sec:exp:setup}.}
    \vspace{-0.5cm} 
  \label{fig:first-page}
  \figurespace{}
\end{figure}

Some large-scale ML tasks are particularly hard to distribute because standard
PM approaches are infeasible or inefficient. Currently, the most widely used
approach is full static replication, i.e., to maintain a copy of the entire
model at each cluster node. Static full replication is infeasible when the model
size exceeds the memory capacity of a single node. It is also inefficient when
the task accesses model parameters \emph{sparsely}, i.e., when each update step
reads and writes only a small subset of all parameters. This is because it
synchronizes updates for all parameters to all nodes, even though each node
accesses only a small subset at each point in time. Such sparse access is common
in natural language processing~\cite{word2vec, glove, elmo, bert}, knowledge
graph embeddings~\cite{complex, rescal, tucker, transe}, some graph neural
networks~\cite{rgcn, wgcn, compgcn}, click-through-rate
prediction~\cite{widedeep, deepfm, deepcross, deepinterest}, and recommender
systems~\cite{mf-recsys, music-rating-prediction,
  collab-filtering-implicit-feedback}. \footnote{\add{For example, sparsity may
    arise due to sparse features (e.g., binary features or embedding layers of
    tokens, entities, or vertices) or due to sampling (e.g., negative sampling
    for classification or neighborhood sampling in GNNs).}} Another standard PM
approach---static parameter partitioning---partitions model parameters and transfers
parameters to where they are needed on demand. This approach is often
inefficient because of the access latency that this ad-hoc transfer
induces~\cite{lapse}. Figure~\ref{fig:first-page} shows that the performance of
both these approaches (blue and red lines) falls behind that of a single node
baseline for an example ML task, thus necessitating more advanced PM.

Advanced PM approaches can improve efficiency \cite{ssp,essp,lapse,bips},
but to do so, they require information about the underlying ML task. However, in
current ML systems, this information is not readily available to the parameter manager.
To work around this lack of information, existing advanced PM approaches require
application developers to control critical PM decisions manually, by exposing
configuration choices and tuning knobs. This requirement makes these PM
approaches complex to use, such that their adoption in common ML systems remains
limited. For example, multi-technique parameter managers \cite{bips,nups} use
different PM techniques (e.g., replication and dynamic allocation) for different
parameters. This can be more efficient than simple PMs (e.g.,~see the
performance of NuPS in Figure~\ref{fig:first-page}), but application developers
need to specify upfront which technique to use for which parameter, and---for
optimal performance---need to tune these technique choices. The absence of a
general-purpose approach that is both simple to use and efficient gives rise to a
range of custom, task-specific approaches that tightly couple PM to one specific
ML task or one class of ML tasks \cite{biggraph, aligraph, paddlebox, fae, het,
  sancus}. These approaches can be efficient, but they are not general-purpose
leading to similar techniques being re-developed and re-implemented many times.

In this paper, we propose \emph{intent signaling}, a general-purpose mechanism
for providing the information that is necessary to enable tuning-free advanced
PM \add{for training with (some) sparse parameter access}. Intent signaling
cleanly decouples the application from PM: the application passes information
about which parameters it \emph{intends} to access before it does so, and the
parameter manager acts based on these signals. We designed intent signaling in a
way that can naturally be integrated into existing ML systems: the data loader
can signal intent during batch preparation. We show that intent signaling not
only simplifies the use of existing advanced PM approaches, it also enables a
range of novel, more fine-grained, parameter managers that so far were
impractical because PM had to be controlled manually.

In addition, we develop \sys{}, a proof-of-concept parameter manager that is based
on intent signaling. \sys{} is (i) \emph{fully adaptive}, i.e., it dynamically
adapts all critical decisions to the underlying ML task, and (ii)
\emph{zero-tuning}, i.e., it adapts automatically, based purely on intent
signals. To do so, \sys{} continuously re-evaluates what is the most efficient
way to manage a specific parameter in the current situation (e.g.,~whether
relocating the parameter is more efficient than replicating it). As the timing
of PM actions (e.g., when to start maintaining a replica at a node)
significantly affects performance, \sys{} continuously optimizes its
action timing by learning from past timing decisions. \sys{} currently provides
automatic PM among the main memory of cluster nodes.\footnote{Workers may, of
  course, use GPUs to process batches. Direct support for GPU memory in \sys{}
  may provide further efficiency improvements and is a key direction for future
  work. }

We conducted an experimental evaluation on five ML tasks: training knowledge
graph embeddings, word vectors, matrix factorization, click-through-rate
prediction models, and graph neural networks. \sys{} was efficient across all
tasks without requiring any tuning. In addition, it matched or even outperformed
existing, manually tuned PM approaches. Figure~\ref{fig:first-page} shows one
example of this performance: \sys{} out of the box matched the performance of
the best NuPS configuration, which was tuned specifically for this ML task.

In summary, our contributions are: (i) we analyze the complexity of existing PM approaches
(Section~\ref{sec:parameter-management-approaches}), (ii) we propose intent
signaling (Section~\ref{sec:intent-signaling}), (iii) we describe \sys{}
(Section~\ref{sec:sys}), and (iv) we investigate the performance of \sys{} in an
experimental study (Section~\ref{sec:experiments}).

\newcommand{\yes}{\textcolor{Worker2}{\textbf{+}}}
\newcommand{\no}{\textcolor{red}{\textbf{-}}}
\newcommand{\syes}{\yes \yes}
\newcommand{\sno}{\no \no}
\newcommand{\ok}{-}

\begin{table*}
  \caption{Approaches to distributed PM: adaptivity, ease of use, and efficiency
    for sparse workloads. Existing approaches adapt only individual system
    aspects to the underlying ML task and require applications to manually
    control these adaptations.}
  \label{tab:approaches-to-parameter-management}
  \centering

  {
  \begin{threeparttable}
  \begin{tabular}{lcccccc} \toprule
    & \multicolumn{4}{c}{Adaptivity}  & & \\
    \cmidrule(lr){2-5}
    Approach & Replication & \hspace{-0.3cm}\makecell{Parameter location}\hspace{-0.2cm} & \makecell{Choice of technique}\hspace{-0.4cm}  & Timing & \hspace{-0.25cm}\makecell{Ease of use}& \hspace{-0.2cm}Efficiency \\
    \midrule
    Static full replication        & static (full)      & static             & single & none            & \syes & \sno  \\
    Static parameter partitioning (PS-Lite)\hspace{-0.3cm}& none               & static             & single & none            & \syes & \sno  \\
    Selective replication (Petuum)    & adaptive           & static             & single & by application\hspace{-0.2cm} & \no   & \no   \\
    Dynamic allocation (Lapse)   & none               & adaptive           & single & by application\hspace{-0.2cm} & \sno  & \no   \\
    Multi-technique PM (BiPS, Parallax)\hspace{-0.2cm} &\hspace{-0.3cm} \makecell{static (partial)}\hspace{-0.2cm}   & static  & static & none             & \sno  & \yes  \\
    Multi-technique PM (NuPS) & \hspace{-0.3cm}\makecell{static (partial)}\hspace{-0.2cm}   & adaptive           & static  & by application\hspace{-0.2cm}  & \sno  & \yes  \\
    \sys{} (this paper)& adaptive           & adaptive           & adaptive & adaptive          & \yes  & \syes \\
    \bottomrule
  \end{tabular}
  \end{threeparttable}}
\end{table*}


\section{Existing Approaches}
\label{sec:parameter-management-approaches}

Several general-purpose PM approaches aim to improve efficiency by adapting
specific aspects to the underlying ML task. This adaptation requires
information about the task. As this information is not available to the
parameter manager in current ML systems, existing approaches rely on the
application developers to manually control
adaptation, which makes these approaches complex to use. In this section, we
briefly discuss existing approaches, which aspects they adapt, and what makes them
complex to use. Table~\ref{tab:approaches-to-parameter-management} gives an
overview. 
A more detailed analysis that also discusses efficiency can be found in 
\iflong
Appendix~\ref{sec:parameter-management-approaches-long}.
\fi%

We discussed static full replication and static parameter partitioning in
Section~\ref{sec:intro}. These are implemented in many ML systems.
For example, TensorFlow~\cite{tensorflow} implements the two in the \emph{mirrored} and
\emph{parameter server} distribution strategies, respectively; PyTorch~\cite{pytorch}
implements them in the \emph{distributed data parallel} and \emph{fully sharded
  data parallel} wrappers, respectively. They are easy to use, but their
efficiency is limited, as discussed in Section~\ref{sec:intro}.

\textbf{Selective replication}~\cite{petuum} statically partitions parameters,
and, to improve efficiency, selectively replicates
parameters to further nodes during training.
Existing approaches~\cite{ssp, essp, WeiDQHCGGGX15} are complex to use because they require
applications to manually tune a staleness threshold parameter that affects both model
quality and run time efficiency for each ML task.

\textbf{Dynamic parameter allocation}~\cite{lapse} partitions parameters to
nodes and, to improve efficiency, dynamically relocates parameters to the nodes
that access them during training.
Existing approaches are complex to
use because they require the application to initiate parameter relocations
manually in the application code and to tune the \emph{relocation offset}
(i.e., how long before an actual access the relocation is initiated).

\textbf{Multi-technique parameter managers} \cite{parallax, bips, nups} support
multiple PM techniques (e.g., replication, static partitioning, or dynamic
allocation) and use a suitable one for each parameter. They are complex to use
because they require the application to specify upfront which technique to use
for which parameter. For optimal performance, application developers
additionally need to tune these technique choices manually.




\section{Intent Signaling}
\label{sec:intent-signaling}

To enable automatic adaptive PM, we propose \emph{intent signaling}, a novel
mechanism that naturally integrates into ML systems. It passes
information about upcoming parameter access from the application to the
parameter manager. It \emph{decouples} information from action, with a clean API
in between: the application provides information (intent signals); the parameter
manager transparently adapts to the workload based on the intent signals. In
other words, all PM-related decision-making and knob tuning is done
transparently by the parameter manager. The application \emph{only signals
  intent}.

\add{An intent is a declaration by one worker that this worker intends to access
  a specific set of parameters in a specific \add{(logical)} time window in the
  future. A natural choice of time window is a training batch. For example, a
  worker may signal that it will require access to, say, only 1M out of 100M
  parameters in batch~2 after analyzing the examples in that batch (e.g.,
  embeddings of categories that do not appear in the batch are not
  needed).\footnote{For example, in the click through-rate prediction task of
    Sec.~\ref{sec:experiments}, processing a batch size with 10000 examples
    requires accesses to less than 0.7\% of all parameters on average.} Such an
  approach integrates naturally with the data loader paradigm of common ML
  systems --- such as Pytorch's data loader~\cite{pytorch}, TensorFlow's data
  sets~\cite{tensorflow}, or the Gluon data loader~\cite{mxnet} ---, where
  training batches are constructed upfront in one or more separate threads and
  then queued until processed by the training thread(s) later on. This process
  is illustrated in Figure~\ref{fig:data-loader}. }

In general, it is important that intent is signaled before the parameter is
actually accessed, such that the parameter manager can adapt proactively. We use
logical clocks as a general-purpose construct for specifying the start and end
points of intents. To ensure generality, each worker~$i$ has one logical
clock~$C^i$ that is independent of other workers' clocks.\footnote{Applications
  may, of course, synchronize worker clocks.} Each worker advances its clock
with an \texttt{advanceClock()} primitive (as done in Petuum~\cite{petuum}, but,
in contrast to Petuum, invocation of our \texttt{advanceClock()} is cheap, as it
only raises the clock). For example, a worker could advance its clock whenever
it starts processing a new batch.

\def \prepsep {1.7}
\def \isep    {3.5}
\def \isepX   {2.8}
\def \asep    {-1.3}

\newcommand\advanceclock[2]{
  \draw[very thick] (#2,-1.7) -- ++(0,-0.4);
  \node at (#2, -2.5) {#1};
}

\begin{figure}

  \centering

\resizebox{1.0\columnwidth}{!}{

  \begin{tikzpicture}[auto,font={\fontsize{13pt}{14}\selectfont},
  prep/.style={align=center, anchor=west, minimum width=1.6cm, minimum height=1.3cm, draw=black, fill=neutralbg, rounded corners=1.5mm},
  batch/.style={align=center, anchor=west, minimum width=1.8cm, minimum height=1.3cm, draw=black, fill=mypurple!20, rounded corners=1.5mm},
  accesses/.style={align=center, anchor=west,color=mypurple!70, text width=1.7cm},
  intentcall/.style={anchor=south west, rotate=18},
  access/.style={->, very thick,>=stealth'},
  intent/.style={->, ultra thick,>=stealth',myorange},
  annot/.style={midway,sloped,above},
  flipannot/.style={midway,sloped,below,rotate=180},
  arrow/.style={->, >=stealth', shorten <= -4pt, shorten >=-2pt, densely dotted},
  ]

  \node [align=center] at (-3, \prepsep) {data loader\\thread(s): };

  \node (prep1) [prep] at (-1.6, \prepsep) {prepare\\batch 1};
  \node (intent1) [intentcall] at ([yshift=0.1cm]prep1.60) {$\mathtt{Intent(\mathcal{P}_1,1,2)}$};
  \draw [arrow] (prep1.70) -- ([yshift=0.1cm]intent1.south west);

  \node (prep2) [prep] at (0.2, \prepsep) {prepare\\batch 2};
  \node (intent2) [intentcall] at ([yshift=0.1cm]prep2.60) {$\mathtt{Intent(\mathcal{P}_2,2,3)}$};
  \draw [arrow] (prep2.70) -- ([yshift=0.1cm]intent2.south west);

  \node (prep3) [prep] at (2, \prepsep) {prepare\\batch 3};
  \node (intent3) [intentcall] at ([yshift=0.1cm]prep3.60) {$\mathtt{Intent(\mathcal{P}_3,3,4)}$};
  \draw [arrow] (prep3.70) -- ([yshift=0.1cm]intent3.south west);

  \node (prep4) [prep] at (4, \prepsep) {prepare\\batch 4};
  \node (intent4) [intentcall] at ([yshift=0.1cm]prep4.60) {$\mathtt{Intent(\mathcal{P}_4,4,5)}$};
  \draw [arrow] (prep4.70) -- ([yshift=0.1cm]intent4.south west);

  \node at (6.2, \prepsep) {\textbf{...}};

  \node [align=center] at (-3, 0) {worker\\thread: };
  \node (tr1) [batch] at (2, 0) {train\\batch 1};
  \node (access1) [accesses] at (2, \asep) {access\\$\mathcal{P}_1$};
  \node (tr2) [batch] at (4, 0) {train\\batch 2};
  \node (access2) [accesses] at (4, \asep) {access\\$\mathcal{P}_2$};
  \node (tr3) [batch] at (6, 0) {train\\batch 3};
  \node (access3) [accesses] at (6, \asep) {access\\$\mathcal{P}_3$};
  \node (tr4) [batch] at (8, 0) {train\\batch 4};
  \node (access4) [accesses] at (8, \asep) {access\\$\mathcal{P}_4$};
  \node at (10.4, 0) {\textbf{...}};

  \draw [->, very thick, >=stealth'] (-1.6, -1.9) -- ++(12.5,0);
  \node at (9.8,-2.5) {time (clock)};

  \advanceclock{0}{-1.6};
  \advanceclock{1}{2};
  \advanceclock{2}{4};
  \advanceclock{3}{6};
  \advanceclock{4}{8};

\end{tikzpicture}

}

\caption{The data loader is a natural place to integrate intent signaling.
  After a batch~$i$ is constructed, the data loader signals intent for the
  parameters $\mathcal{P}_i$ that are accessed in batch~$i$. }
\label{fig:data-loader}
\figurespace{}

\end{figure}
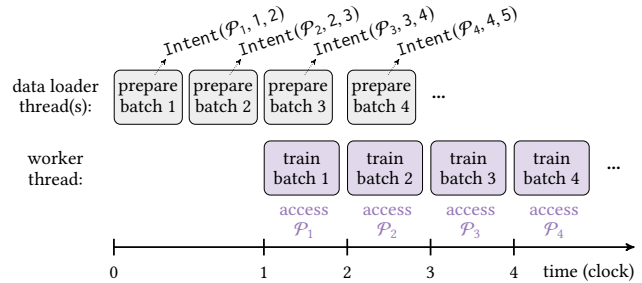

We propose the following primitive for signaling intent:
\[
  \mathtt{Intent(\ parameters,\ \ \ C_{start}, \ \ \ C_{end}\ \ \  [,  type] \ )}%
\]
With this primitive, a worker signals that it intends to access a set of
parameters in the time window between a start clock $\mathtt{C_{start}}$
(inclusive) and an end clock $\mathtt{C_{end}}$ (exclusive). The primitive
allows to (optionally) specify intent type, e.g., \emph{read}, \emph{write},
\emph{read+write}. Figure~\ref{fig:data-loader} illustrates an example of how
the data loader can signal intent. With $\mathtt{Intent(\mathcal{P}_2, 2, 3)}$, the data
loader signals that the corresponding worker intends to access parameters
$\mathtt{\mathcal{P}_2}$ (e.g., 1M out of 100M parameters) when at clock~2 (i.e., while it
trains on batch~2). We say that an intent is \emph{inactive} if it is signaled,
but the worker has not reached the start clock yet, i.e., $\mathtt{C^i <
  C_{start}}$. We say that an intent is \emph{active} if the worker clock is
within the intent time window, i.e., $\mathtt{C_{start} \leq C^i < C_{end}}$. And
we say that an intent is \emph{expired} when the worker clock has reached the
end clock, i.e., when $\mathtt{C_{end} \leq C^i}$. Invocation of the
$\mathtt{Intent}$ primitive is meant to be cheap, i.e., it should not slow down
the worker, even if the worker signals a large number of intents. Workers can
flexibly combine intents: they can signal multiple (potentially overlapping)
intents for the same parameter, extend one intent by signaling another
later on, etc.

\add{Intent signaling enables the PM to continuously adapt its parameter
  management strategy. This is in contrast to most existing PM approaches, where
  applications directly or indirectly control which or when actions such as
  selective replication are performed. As we will discuss in
  Section~\ref{sec:sys}, intent signals allow \sys{} to choose suitable PM
  techniques dynamically during run time (as opposed to statically per
  parameter) and to time actions appropriately (as opposed to explicitly
  triggering actions). The key benefits of intent signaling are (i) ease of use,
  as applications do not need to make these hard choices (but only signal
  intent), and (ii) improved efficiency, as the PM can make better choices by
  carefully deciding which and when to perform actions.}


\section{The \sys{} Parameter Manager}
\label{sec:sys}

Intent signaling opens a large design space for adaptive PM. Key
design questions include: when and where to
maintain replicas, whether and when to change parameter allocation, when to act on intent signals, how to synchronize replicas,
how to exchange intent signals, on which nodes to make decisions, and how to
communicate efficiently. We explore this space and describe
\sys{}, a parameter manager that---in contrast to previous
ones---is \emph{fully adaptive}, i.e., it adapts \emph{all} critical aspects
to the underlying ML task, see
Table~\ref{tab:approaches-to-parameter-management}. It does so
\emph{automatically}, based purely on intent signals, thus requiring no user
input or knob tuning. And it does so \emph{dynamically}, continuously re-evaluating
what the most efficient approach is for the current situation. 
\add{AdaPM has low overhead ---a major design goal---and it takes all management decisions off the critical path of workers.}

We give a brief overview of key design choices before we discuss the most
challenging ones in detail. Figure~\ref{fig:architecture} illustrates the
architecture of \sys{}. We assume an architecture that---for efficiency---co-locates
the parameter store in the same processes as the worker threads, as parameter
managers commonly do~\cite{ssp, flexps, lapse}.

\definecolor{neutralbg}{HTML}{eeeeee}

\tikzset{	
	col_arrow_style/.style={<->,>=stealth',line width=0.2mm, color=black, shorten <=3pt, shorten >=3pt}, 
	col_rectangles/.style={draw=black,fill=mypurple!30, rounded
    corners=0.7mm,minimum height=5.0mm,minimum width=1.75cm, anchor=north west,
    yshift=-0.5mm, xshift=0mm, align=center, text centered},
	col_small_arrows/.style={<->,>=stealth', line width=0.26mm, shorten <=2pt, shorten >=2pt},
  store/.style={minimum width=1.2cm, anchor=north east, align=center, font=\large}
}

\def \colfontsize {\large}

\newcommand\showtext[1]{#1}

\newcommand{\archworker}[7]{
	\node [minimum width=4.70cm, minimum height = 3.35cm,draw=black, rounded corners, fill=neutralbg, align=center,scale=#6](nd#1) [label=below:{\colfontsize #2}] {};


	\node (nd#1a) [scale=#6,col_rectangles, anchor=north west, xshift=1.5mm, yshift=-1.0mm] at (nd#1.north west) {\showtext{\colfontsize{worker 1}}};

	\node (nd#1b) [scale=#6, col_rectangles] at (nd#1a.south west) {\showtext{\colfontsize{worker 2}}};

	\node (nd#1c) [scale=#6, col_rectangles,draw=none,fill=none,minimum height=0mm,yshift=+0.6mm] at (nd#1b.south west) {...};

	\node (nd#1d) [scale=#6, col_rectangles,yshift=+0.6mm] at (nd#1c.south west) {\showtext{\colfontsize{worker n}}};

	\node (nd#1e) [scale=#6, col_rectangles,fill=mypurple!20] at (nd#1d.south west) {\showtext{\colfontsize{server}}};
	\node (nd#1f) [scale=#6, col_rectangles,fill=mypurple!20] at (nd#1e.south west) {\showtext{\colfontsize{sync.}}};

  \begin{scope}[shift={(0.68*#6,-0.57*#6)},scale=0.27*#6,name prefix=parameters]
    \foreach \i in {0,...,17}
    \baseP{\i}{0}{0}
    ;

    #7

    \draw [thick] (0,0) rectangle ++(3,6);
    \node (frame) [draw=none,fill=none] at (1.5, 0) {}; 
  \end{scope}

	\node (nd#1intents) [store, scale=#6, yshift=-5.5,
  xshift=0mm,fill=third!50,draw,minimum
  height=0.4cm,anchor=north,minimum width=1.19cm] at (parametersframe) {\showtext{intents}};

	\draw [scale=#6,col_small_arrows, shorten <=0.1cm, shorten >=0.1cm]
  (nd#1c.east) to  ([xshift=1.1cm]nd#1c.east) node[midway, align=center, font=\large,
  anchor=center, xshift=+0.24cm, yshift=0.8cm,scale=#6] {\showtext{shared\\mem.}};

}

\begin{figure}
  \centering

  \begin{tikzpicture}[
   	line width = 0.2mm,
    scale=0.70,
    every node/.style={scale=0.70}
    ]

    \matrix[row sep=-9mm,column sep=2mm]  {
      & \archworker{1}{process at node 1}{1.2cm}{1.0}{}{1.0}{
        \repPsingle{17}{0}{0};
        \allocP{2}{0}{0};
        \allocP{1}{0}{0};
        \allocP{3}{0}{0};
        \allocP{7}{0}{0};
        \allocP{10}{0}{0};
        \allocP{13}{0}{0};
      }; &  \\
      \renewcommand\showtext[1]{}
      \archworker{2}{process at node 2}{0.8cm}{0.8}{[-3pt]}{0.6}{
        \repPsingle{2}{0}{0};
        \repPsingle{5}{0}{0};
        \allocP{0}{0}{0};
        \allocP{6}{0}{0};
        \allocP{8}{0}{0};
        \allocP{12}{0}{0};
        \allocP{14}{0}{0};
        \allocP{16}{0}{0};

      }; & & \renewcommand\showtext[1]{}\archworker{3}{process at node 3}{1.7cm}{1.16}{}{0.6}{
        \allocP{17}{0}{0};
        \allocP{4}{0}{0};
        \allocP{5}{0}{0};
        \allocP{9}{0}{0};
        \allocP{11}{0}{0};
        \allocP{15}{0}{0};
      };
      \\};

    \begin{scope}[shift={(-5.5,0.48)},scale=0.27,name prefix=parameters]

      \draw [fill=gray!10,draw=none] (-1.0,-1.0) rectangle (10.2, 8);

      \node [anchor=west] at (-0.5, 6.5) {Parameters};

      \baseP{12}{0}{0};
      \node [anchor=west] at (1.6, 4.5) {no local access};

      \allocP{6}{0}{0};
      \node [anchor=west] at (1.6, 2.5) {main copy};

      \repPsingle{0}{0}{0};
      \node [anchor=west] at (1.6, 0.5) {replica};

    \end{scope}

    \node at (5.5,0) {};

    \draw [col_arrow_style]([yshift=-5mm]nd2.east) to [bend right=10] node[above] {} ([yshift=-5mm]nd3.west);
    \draw [col_arrow_style](nd3.north) to [bend right=30] node[below,xshift=8mm,yshift=13mm,align=center,font=\large] {} (nd1.east);
    \draw [col_arrow_style](nd1.west) to [bend right=30] node[below] {} (nd2.north);
    
  \end{tikzpicture}

  \caption{\sys{} architecture. For efficiency, \sys{} runs multiple worker
    threads in one process per node, and accesses locally available parameters
    via shared memory.}
  \label{fig:architecture}
  \figurespace{}
\end{figure}
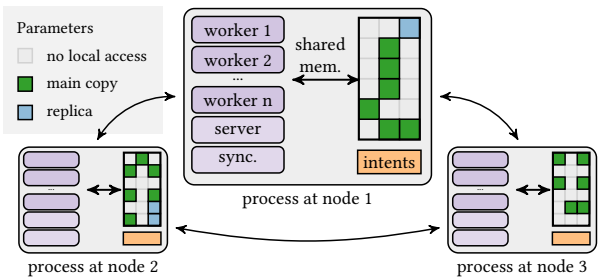


\textbf{Adaptive choice of technique. } \sys{} employs dynamic allocation and
selective replication, and adaptively picks between the two per parameter.
\sys{} dynamically chooses the most efficient technique for the current
situation. Intuitively, \sys{} relocates a parameter if---at one point in
time---only one node accesses the parameter. Otherwise, it creates replicas
precisely where they are needed. We discuss \sys{}'s choice of
technique in Section~\ref{sec:sys:relocation-and-selective-replication}.

\textbf{Adaptive action timing. } \sys{} learns automatically when the right
time to act on an intent signal is. This ensures that applications do not need to
fine-tune the timing of their intent signals. They can simply
signal their intents early, without sacrificing performance. See
Section~\ref{sec:sys:timing} for details.

\textbf{Responsibility follows allocation.} In \sys{}, the node that currently
holds the main copy of a parameter (owner node) takes on the main responsibility
for managing this parameter: it decides how to act on intent signals and acts as
a hub for replica synchronization. For efficiency, this responsibility moves
with the parameter whenever the it is relocated. \add{This strategy generally
  reduces network communication as the responsibility for a parameter is always
  at a node that ``needs'' it.}
\iflong
Details can be found in Section~\ref{sec:sys:owner-node} (page~\pageref{sec:sys:owner-node} of the appendix).
\fi

\newcommand{\fsx}[0]{\Large}

\begin{figure*}[t]

    \policyplotbase{Worker2}
    \node at (-1,-1.3) {}; 
    \draw [fill=gray!10,draw=none] (6.0,-0.5) rectangle (55.5, 1.5);
    \intentsignal{0}{7.5}{1.5}
    \node [anchor=west] at (8.1,0.5) {\small intent signal};
    \actionwindow{0}{15}{1.5}
    \node [anchor=west] at (16.4,0.5) {\small action window};
    \intent{0}{25}{1.5}
    \node [anchor=west] at (26.4,0.5) {\small active intent};
    \alloc{0}{33.5}{1.5}
    \node [anchor=west] at (35.0,0.5) {\small parameter allocation};
    \rep{0}{45.5}{1.5}
    \node [anchor=west] at (46.9,0.5) {\small parameter replica};

    \policyplotend{}

    \begin{subfigure}[c]{.19\textwidth}
      \centering

      \vspace{-0.69cm}
      \resizebox{0.8\columnwidth}{!}{
        \begin{tikzpicture} []
          \node (decision) [diamond,fill=lightgray,align=center,inner sep=0mm,minimum size=3.5cm] at (0, 0) {};

          \node at (decision) [align=center] {\fsx{} Num.\\\fsx{}nodes with\\\fsx{} active intent\\\fsx{}at time\\\fsx{}$t$?};

          \node (relocate) [fill=Worker2!80,align=center,inner sep=2mm] at (2.2,-1.4) {\fsx{}Relocate to \\\fsx{} the node\\\fsx{}with intent};
          \node (replicate) [fill=Worker1!50,align=center,inner sep=2mm] at (0.9,-2.9) {\fsx{}Maintain replicas on the\\\fsx{}nodes with active intent};

          \draw[very thick, ->, >=stealth'] (decision.east) to [out=0, in=90] node[midway,above,yshift=0.0cm,xshift=0.1cm] {\Large\textbf{1}} (relocate.north);
          \draw[very thick, ->, >=stealth'] (decision.south) to [out=270,in=110] node[midway,yshift=0.10cm,xshift=-0.6cm] {\Large\textbf{>1}} (replicate.140);
        \end{tikzpicture}
      }

      \caption{Decision}
      \label{fig:decision}
    \end{subfigure}%
  \begin{subfigure}[c]{.27\textwidth}
    \centering

    \adapolicy{Worker2}
    \alloc{0}{0}{2}
    \alloc{2}{2}{6}
    \alloc{3}{8}{4}

    \signal{2}{0.5}{2.5}
    \intent{2}{3}{2}

    \signal{3}{4}{6}
    \intent{3}{9}{1}
    \policyplotend{}

    \caption{Non-overlapping intents}
    \label{fig:scenario:separate}
  \end{subfigure}%
  \begin{subfigure}[c]{.27\textwidth}
    \centering

    \adapolicy{Worker2}
    \alloc{0}{0}{2}
    \alloc{2}{2}{6}
    \alloc{3}{8}{4}
    \preprep{3}{6}{2}

    \signal{2}{0.5}{2.5}
    \intent{2}{3}{5}

    \signal{3}{3}{3}
    \intent{3}{6}{4}
    \policyplotend{}

    \caption{Partially overlapping intents}
    \label{fig:scenario:overlap}
  \end{subfigure}%
  \begin{subfigure}[c]{.27\textwidth}
    \centering

    \adapolicy{Worker2}
    \alloc{0}{0}{10}
    \preprep{0}{11.8}{0.2}

    \preprep{1}{1}{4}
    \preprep{1}{6}{2}
    \preprep{1}{9}{1}
    \alloc{1}{10}{1}
    \preprep{1}{11.9}{0.1}

    \preprep{2}{1}{3}
    \preprep{2}{5}{4}
    \preprep{2}{11}{0}
    \alloc{2}{11}{1}

    \preprep{3}{6}{4}

    \customsigintent{0}{11.8}{0.2}{2.1}
    \sigintent{0}{3.8}{5}
    \intent{0}{0}{3}

    \sigintent{1}{11.9}{0.1}
    \sigintent{1}{9}{2}
    \sigintent{1}{6}{2}
    \customsigintent{1}{1}{4}{1.0}

    \sigintent{2}{11}{1}
    \sigintent{2}{5}{4}
    \customsigintent{2}{1}{3}{1.1}

    \customsigintent{3}{6}{4}{2.2}
    \policyplotend{}

    \caption{Many concurrent intents (hotspot)}
    \label{fig:scenario:hotspot}
  \end{subfigure}%
  \caption{\sys{} decides automatically whether to relocate or replicate a
    parameter at any time $t$ (a). Example scenarios (b--d).  }
  \label{fig:policies}
  \figurespace{}
\end{figure*}
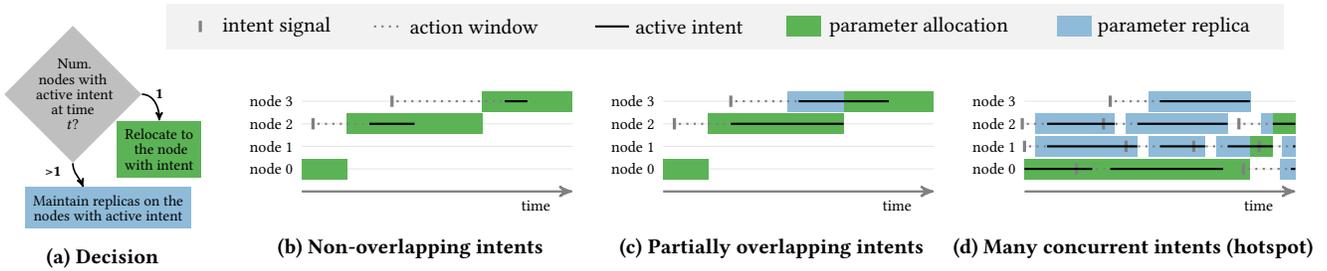


\textbf{Efficient communication. } \sys{} communicates for exchanging intent
signals, relocating parameters, and managing replicas. To communicate
efficiently, \sys{} locally aggregates intents,
groups messages when possible to avoid small message overhead,
and employs location caches to improve routing. 
\iflong
Details can be found in Section~\ref{sec:sys:efficient-communication} (page~\pageref{sec:sys:efficient-communication} of the appendix).
\fi

\textbf{Optional intent. } Intent signals are optional in \sys{}. An
application can access any parameter at any time, without signaling intent.
However, signaling intent makes access more efficient, as it allows
\sys{} to avoid synchronous network communication.

\subsection{Adaptive Choice of Technique}
\label{sec:sys:relocation-and-selective-replication}

\sys{} receives intent signals from workers. Based on these intent signals,
\sys{} tries to ensure that a parameter can be accessed locally at a node
while this node has active intent for this parameter. To achieve this, \sys{} has
to determine where to ideally allocate a parameter, if and where to create
replicas, and for how long to maintain each replica.

To make parameters available locally, \sys{} employs (i) dynamic allocation and
(ii) selective replication. That is, it (i) can relocate parameters among nodes
and (ii) can selectively create replicas on subsets of nodes for specific time
periods. \sys{} uses a simple heuristic rule to decide between the two: if, at
one point in time, only \emph{one} node has active intent for a parameter,
\sys{} relocates the parameter to the node with active intent. After the node's
intent expires, \sys{} keeps the parameter where it is until some other node
signals intent. In contrast, when multiple nodes have active intent for one
parameter concurrently, \sys{} selectively creates a replica at each of the
nodes when the intent of that node becomes active. It destructs the replica when
the intent of that node expires. The simplicity of the decision rule allows
\sys{} to communicate intent among nodes efficiently. Figure~\ref{fig:decision}
illustrates \sys{}'s decision between relocation and selective replication.
\add{More involved approaches (or cost models) may provide further benefits, but
  this simple strategy performed remarkably well in our experimental study.}


Let us consider three exemplary intent scenarios:

\begin{enumerate}
\item Two nodes have intent for the same parameter, and the active phases of the
  intents do not overlap; see Figure~\ref{fig:scenario:separate}. \sys{}
  relocates the parameter from its initial allocation to the first node with
  intent and keeps it there even after the intent expires. \sys{} relocates the
  parameter to the second node with intent shortly before the node's intent becomes
  active.

\item Two nodes have intent for the same parameter, and the active periods of
  the intents partially
  overlap; see Figure~\ref{fig:scenario:overlap}. \sys{} relocates the
  parameter to the first node with intent, then creates a replica on the second
  node while the two active intents overlap, and finally relocates the parameter to the
  second node after the intent of the first node expires.

\item Multiple nodes repeatedly have intent for the same parameter, see
  Figure~\ref{fig:scenario:hotspot}. \sys{} creates replicas on all nodes with
  active intent. Whenever there is exactly one node with active intent (and the
  parameter is not currently allocated at this node), \sys{} relocates the
  parameter to this node.
\end{enumerate}

\sys{} combines parameter relocation and replication because previous research
has shown that their combination is beneficial~\cite{nups}. Relocation is
efficient for parameters that are accessed infrequently, as in the first example
above, because the parameter value is transferred over the network only once per
access (from where the parameter currently is to where the intent is). In
contrast, replication can significantly reduce network overhead for frequently
accessed parameters. In addition, unlike previous multi-technique
approaches~\cite{nups, bips}, \sys{} employs \emph{selective} replication:
\sys{} creates a replica precisely for the time during which it is needed. This
increases efficiency as \sys{} does not need to maintain replicas while they are
not needed. This also limits the impact of stale reads --- which always arise when
replication is used --- since (i) AdaPM replicates as few parameters and as
selectively as possible and continuously refreshes them
(cf.~Tab.~\ref{tab:comm-and-freshness}) and (ii) no staleness occurs for reads
of any non-replicated parameter. Further, unlike previous approaches, the choice
of PM techniques for each parameter is dynamic: 
\sys{} can relocate the parameter at one point in time, and replicate it at
another.

For making its decisions, \sys{} treats all intent types identically. More
complex approaches could tailor their choices to intent type. For
example, systems could choose to take different actions for \emph{read} and
\emph{write} intents. In \sys{}, we keep the system simple and treat all intent
types identically because we do not expect tailoring to improve performance
for typical ML workloads: (i) applications typically both read \emph{and} write
a parameter and (ii) synchronous remote reads are so expensive that it is
beneficial to provide a locally accessible value for a parameter even for a single read.

\subsection{Adaptive Action Timing}
\label{sec:sys:timing}

\sys{} receives intent signals before the intents become active. I.e., there
is an \emph{action window} between the time the intent is signaled and the time
the intent becomes active. \sys{} needs to determine at which point in this
action window it should start to act on the intent signal, i.e., when it relocates the
parameter or sets up a replica for this parameter. For example, consider the
intent of node~3 in Figure~\ref{fig:scenario:overlap}: \sys{} needs to figure
out at which point in time it starts maintaining a replica on node~3.

Relocating a parameter or setting up a replica takes some time. Consequently, if
\sys{} acts too late and relocation or replica setup is not finished in time,
remote parameter access is required. On the other hand, if \sys{} acts too
early, it might maintain a replica longer than needed, inducing unnecessary communication.
Furthermore, if \sys{} acts too early, it might use replication in
scenarios in which---with better timing---relocation would have been both possible
and more efficient.
However, acting on an intent signal (slightly) too early is much cheaper than
acting too late because the remote accesses caused by too late action slow down training
significantly.
In contrast, acting slightly too early merely causes over-communication. Thus, it is
better to err on the side of acting too early.

The key challenge is that both (i) the \emph{preparation time}
for relocation or replica setup and (ii)
the length of the action window are unknown. The action window length is
unknown because it is unclear when the worker will reach the intent's start clock.
Both times are affected by many factors, e.g., by the
application, the compute and network hardware, and the
utilization of that hardware.

\subsubsection{Learning When to Act}

\sys{} aims to \emph{learn} when the right time is to act on an intent signal. A
general approach would estimate both preparation time and action window length
separately. However, \sys{} acts on intent signals in point-to-point
communication rounds that take a fairly constant amount of time. \sys{} thus
simplifies the general approach and directly estimates the number of worker
clocks per communication round. This allows \sys{} to decide whether an intent
signal should be included in the current round or if it suffices to include the
signal in a later round. A later round suffices if the \emph{next} round will
finish before the worker reaches the start clock of the intent.

\def\est{\hat\lambda}

As acting (slightly) too early is much cheaper than acting too late, our goal is
to estimate a soft upper bound for the number of clocks during one communication
round. I.e., we want to be
confident that the true number of clocks only rarely (ideally, never)
exceeds this soft upper bound. To this end, we employ a probabilistic approach
in \sys{}: we assume that the number of clocks follows a Poisson distribution,
estimate the (unknown) rate parameter for the distribution from past
communication rounds, and use a high quantile of this Poisson distribution
as a soft upper bound (e.g., the $0.9999$ quantile). In detail, we assume that the
number of clocks by worker~$i$ in round~$t$ follows
$\operatorname{Poisson}(\lambda^i_t)$ with expected rate $\lambda^i_t$. We
choose a Poisson distribution because it is the simplest, most natural
assumption, and it worked well in our experiments. Note that we assume a Poisson
distribution for a short time period (one round~$t$ by one worker~$i$), not
one global distribution (a much stronger, unrealistic assumption, which,
for example, would not account for changes in workload or system load).

\begin{figure}

  \centering
  \resizebox{1.0\columnwidth}{!}{
    \begin{tikzpicture}[>=stealth',auto]
  \node [anchor = south west] at (0,0) {\includegraphics[page=1]{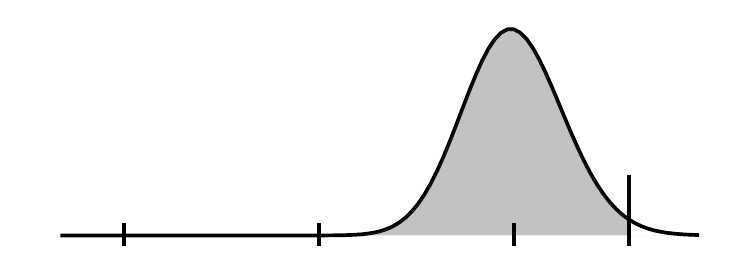}};


  \node (C) at (1.35,0) {$C^i_t$};
  \node (Clab) [anchor=south,align=center,text=gray,scale=0.7] at (1.05,0.65) {current\\worker\\clock};
  \draw [->, gray, shorten >=-4pt] (Clab.south) to [out=270,in=130] (C.north west);

  \draw [decorate, thick, decoration = {brace}] (3.2,0.35) -- +(-1.7,0);
  \node (E1) at (2.35,0) {$\est^i_t$};
  \node (E1lab) [anchor=south,align=center,text=gray,scale=0.7] at (2.05,1.2) {clocks\\during \emph{this}\\comm.~round\\(round $t$)};
  \draw [->, gray, shorten >=-4pt] (E1lab.south) to [out=270,in=110] ([yshift=6]E1.north west);

  \draw [decorate, thick, decoration = {brace}] (5.2,0.35) -- +(-1.7,0);
  \node (E2) at (4.35,0) {$\est^i_t$};
  \node (E2lab) [anchor=south,align=center,text=gray,scale=0.7] at (3.45,0.65) {clocks\\during \emph{next}\\comm.~round\\(round $t+1$)};
  \draw [->, gray, shorten >=-4pt] (E2lab.south) to [out=310,in=130] ([yshift=6,xshift=-6]E2.north west);

  \node at (4.05,2.5) {$\operatorname{Poisson}(2\est^i_t)$};
  \node at (3.85,2.1) {\small (shifted by $C^i_t$)};

  \draw [Worker2,opacity=0.3,thick] (1.375,-0.30) -- +(0,0.9);
  \draw [Worker2,opacity=0.3,thick] (6.5,-0.30) -- +(0,0.9);
  \draw [Worker2, decorate, thick, decoration = {brace}] (6.5,-0.35) -- (1.375,-0.35);
  \node [Worker2, align=center, scale=0.8] at (3.95,-0.75) {In communication round $t$, \sys{} acts on\\intents that start in this window.};

  \node [align=center] at (6.7,1.5) {0.9999\\[-1mm]quantile};

\end{tikzpicture}

}
\caption{\sys{} learns automatically when to act on an intent signal, using a
  probabilistic model to estimate a soft upper bound.}
\label{fig:action-timing-estimate}
\figurespace{}

\end{figure}
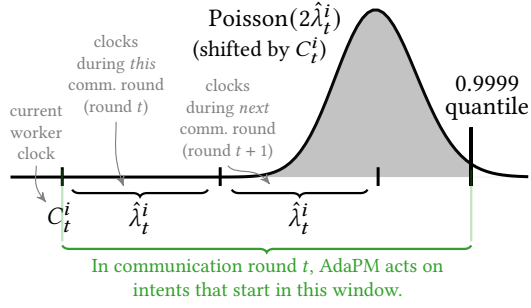


\sys{} acts on a given intent in round $t$ if it estimates that the
corresponding worker might reach the start clock of the intent
($C_{\operatorname{start}}$) before round $t+1$ finishes, i.e.,
roughly\footnote{The exact decision is given in Algorithm~\ref{alg:timing}
  in Section~\ref{sec:aat:rate}.} if
\[
  C_{\operatorname{start}} < C^i_t + Q_{\operatorname{Poiss}}(2 \cdot \lambda^i_t, p)
\]
where $C^i_t$ is the current clock of worker~$i$ at the start of round~$t$ and
$Q_{\operatorname{Poiss}}(\lambda, p)$ computes the $p$ quantile of a Poisson
distribution with rate parameter $\lambda$. Figure~\ref{fig:action-timing-estimate} illustrates
this decision.
Under our Poisson assumption, a $p = 0.9999$ quantile gives a 99.99\% probability that the
actual number of clocks during the two rounds is
below our estimate.

\subsubsection{Estimating the Rate Parameter}
\label{sec:aat:rate}
Naturally, the true Poisson rate $\lambda^i_t$ is unknown. \sys{} estimates this
rate from the number of clocks in past communication rounds, using
exponential smoothing:
\[
  \est^i_{t} \gets (1 - \alpha) \est^i_{t-1} + \alpha (C^i_{t}-C^i_{t-1})
\]
where $\est^i_t$ is the estimate for the number of clocks by worker~$i$ in round
$t$ and $\alpha$ is the smoothing factor.

 \begin{algorithm}[tb]
   \caption{Whether to act on a given intent in round $t$.}
   \label{alg:timing}
   \textbf{Input:} Intent start $C_{\operatorname{start}}$, previous
   estimate $\est^i_{t-1}$, clock of worker $i$ at the start of round $t$
   ($C^i_{t}$) and round $t-1$ ($C^i_{t-1}$), smoothing factor $\alpha$,
   quantile $p$.

   \begin{algorithmic}[1]
     \STATE{$\Delta \gets C^i_{t}-C^i_{t-1}$}

     \IF{$\Delta > 0$\label{line:noupdate-decision}}
     \STATE{
       $\est^i_{t} \gets (1 - \alpha) \est^i_{t-1} + \alpha (\Delta)$
       \;
     }
     \ELSE
     \STATE {
       $\est^i_{t} \gets \est^i_{t-1}$\label{line:noupdate}
       \;
     }
     \ENDIF

     return $C_{\operatorname{start}} < C^i_t + Q_{\operatorname{Poiss}}(2 \cdot
     \operatorname{max}( \est^i_t,  \Delta),
     p)$\label{line:decision}
     \;
     \end{algorithmic}
 \end{algorithm}


We consider two further aspects to improve the robustness of the estimate
$\est^i_t$. First, in ML training tasks, there commonly are periods in which the
workers do not advance their clocks at all. For example, this is commonly the
case at the end of an epoch, while training is paused for model evaluation. In
such periods, the estimate would shrink. To
keep it more constant, \sys{} does not update the
estimate when the worker did not raise its clock during the previous
communication round (i.e., if $C^i_{t}-C^i_{t-1} = 0$).

Second, the observed number of clocks during round $t-1$ (i.e., $C^i_{t}-C^i_{t-1} =
\Delta$) is not independent of the estimate~$\est^i_{t-1}$: if the estimate was
too low, \sys{} did \emph{not} act on some intents that the worker reached in
this round, so that the worker was potentially slowed down drastically by remote parameter
accesses. Thus, the estimate
could settle in a ``slow regime''. A large enough Poisson quantile (i.e., $p \gg
0.5$) ensures that the estimate grows out of such regimes over time.
\sys{} further uses a simple heuristic to get out more
quickly: if the number of clocks in the last round is larger than the current
estimate, it uses this number rather than the estimate (i.e.,
it uses $\operatorname{max}(\est^i_t, \Delta)$).

Algorithm~\ref{alg:timing} depicts precisely how \sys{} decides whether to act
on a given intent and how it updates the estimate.

\subsubsection{Effect on Usability}

\sys{}'s action timing relieves applications from the need to signal
intent ``at the right time'', as is, for example, required for initiating
relocations in dynamic allocation PM approaches. It is important that
applications signal intent early enough so that there is enough time for \sys{}
to act on the signal. Below this lower limit, however, action timing
makes \sys{} insensitive to when intent is signaled. Thus, applications can
simply signal intent early, and rely on \sys{} to act at the right time.

\sys{}'s adaptive action timing introduces three hyperparameters: the smoothing
factor $\alpha$, the quantile $p$, and the initial rate estimate $\est^i_{0}$.
However, these hyperparameters do not require task-specific tuning. We used the
same configuration for all ML tasks and all experiments ($\alpha = 0.1$, $p =
0.9999$, and $\est^i_{0}=10$). This configuration worked well across all five
tested tasks, see Section~\ref{sec:exp:single}.

\section{Experiments}
\label{sec:experiments}

We conducted an experimental study to investigate whether and to what extent
automatic fully adaptive PM is possible and beneficial. The source code,
datasets, and information for reproducibility are available
online.\footnote{\url{https://github.com/alexrenz/AdaPM/}}

In our study, we evaluated the performance of \sys{} by comparing it to
efficient single-node baselines (Section~\ref{sec:exp:single}), a manually tuned
state-of-the-art parameter manager (Section~\ref{sec:exp:nups}), and standard PM
approaches (Section~\ref{sec:exp:standard}). Further, we investigated whether
and to what extent \sys{} benefits from supporting multiple management
techniques (Sections~\ref{sec:exp:ablation} and~\ref{sec:exp:why-relocation}),
how scalable it is (Section~\ref{sec:exp:scalability}), whether action timing is
crucial for its performance (Section~\ref{sec:exp:action-timing}), how
its performance compares to GPU implementations (Section~\ref{sec:exp:task-specific}).
and which decisions it makes in practice (%
\iflong%
Appendix~\ref{sec:exp:in-action}
\fi%
). 
Our major
insights are: (i) \sys{} provided good speedups out of the box, without any
tuning, (ii) \sys{} matched or even outperformed a manually tuned
state-of-the-art parameter manager, (iii) \sys{} scaled efficiently, and (iv)
adaptive action timing is a key building block for \sys{}'s efficiency. We
conclude that \emph{automatic PM is possible and can be efficient}.

\subsection{Experimental Setup}
\label{sec:exp:setup}

\textbf{Tasks. } We considered five ML tasks: knowledge graph embeddings
(KGE), word vectors (WV), matrix factorization (MF), click-through-rate
prediction (CTR), and graph neural networks (GNN). The tasks differ in multiple
ways, including the size of the models, the size of the data set, with what rate
workers advance their clocks, and in their access patterns. 
\add{In particular, some workloads exhibit a large amount of dense accesses
  (e.g., 52\% for CTR), some very few (e.g., 0\% for MF).} We used common
practices for task training, e.g., for partitioning training data and measuring
model quality. When necessary, we tuned hyperparameters for each task on a
single node and used the best found setting in all systems and variants.
\iflong
Appendix~\ref{sec:task-details}
\fi
describes each task in detail.
\iflong
Table~\ref{tab:datasets} (in the appendix) provides an overview.
\fi

\textbf{Baselines.} We compared the performance of \sys{} to efficient
single-node implementations, to NuPS~\cite{nups} (a state-of-the-art
multi-technique parameter manager), to static parameter partitioning, to static
full replication, and to three ablation variants.
To ensure a fair comparison to NuPS, we ran six
different hyperparameter configurations of NuPS for each task. Five configurations are designed to simulate a typical
hyperparameter search by an application developer: a random
search that is loosely informed by the NuPS heuristic and intuition, 
see 
\iflong
Appendix~\ref{sec:nups-configurations} 
\fi
for details. 
In addition, we ran NuPS
with the tuned hyperparameters from Renz-Wieland et al.~\cite{nups}. These
were tuned in a series of detailed experiments (but for a
setting with 8 worker threads per node, not 32). 
Note that such detailed insights are not commonly available to application developers. 
As a single node baseline, we used an efficient shared memory implementation.

\textbf{Implementation and cluster.} We implemented \sys{} in C++, using ZeroMQ
and Protocol Buffers for communication. We used a cluster of
up to 16 Lenovo ThinkSystem SR630 computers, running Ubuntu Linux 20.04,
connected with 100~Gbit/s Infiniband. Each node was equipped with two Intel Xeon
Silver 4216 16-core CPUs, 512~GB of main memory, and one 2~TB D3-S4610 Intel
SSD. We compiled code with g++ 9.3.0. The CTR and GNN tasks are implemented
using PyTorch 1.12.1 and ran with Python 3.9.13. All other tasks are implemented
in C++. We consistently used 32 worker threads per
node and, unless specified otherwise, 8 cluster nodes. In NuPS and \sys{}, we
additionally used 3 ZeroMQ I/O threads per node. In \sys{}, we used 4
communication channels per node; in NuPS, we used 1 channel as it supports only 1.
\add{As
  prior work, we use asynchronous SGD throughout so workers do not block.
  Thus, reads to non-replicated parameters (e.g., in static partitioning,
  NuPS, \sys{}) are always current. Reads to replicated parameters, however, may
  be stale (e.g., in full replication, NuPS, \sys{}). This staleness is
  generally bounded by the time between refreshs;
  see~\cite{ssp,async-sgd-nonconvex} for a convergence analysis under bounded
  staleness.}

\textbf{Measures.} \add{To keep costs controlled, we ran all variants with a
  fixed \SI{4}{h} time budget. This time budget was sufficient for convergence
  for the fastest methods.} We measured model quality over time and over epochs
within this time budget. We conducted 3 independent runs of each experiment,
each starting from a distinct randomly initialized model, and report the mean.
For NuPS, we ran each of the 6 configurations once. We depict error bars for
model quality and run time; they present the minimum and maximum measurements.
In some experiments, error bars are not clearly visible because of small
variance. Gray shading indicates performance that is dominated by the single
node. We report two types of speedups: (i) \emph{raw speedup} depicts the
speedup in epoch run time, without considering model quality; (ii)
\emph{effective speedup} depicts the improvement w.r.t. time-to-quality. It is
calculated from the time that each variant took to reach 90\% of the best model
quality that we observed in the single node. \add{We chose this rather low
  threshold to determine speed-ups also for slower variants (which otherwise
  would not achieve the threshold in the time budget). The speed-ups for AdaPM
  at a higher accuracy, e.g.,~99\%, are near-identical to the ones at 90\%
  (e.g., for CTR, 6.4x for both levels).}

\begin{figure*}
  \panelfiguresetup{}
  \centering

  \begin{subfigure}[b]{0.33\textwidth}
    \centering
    \includegraphics[page=1,width=0.95\textwidth]{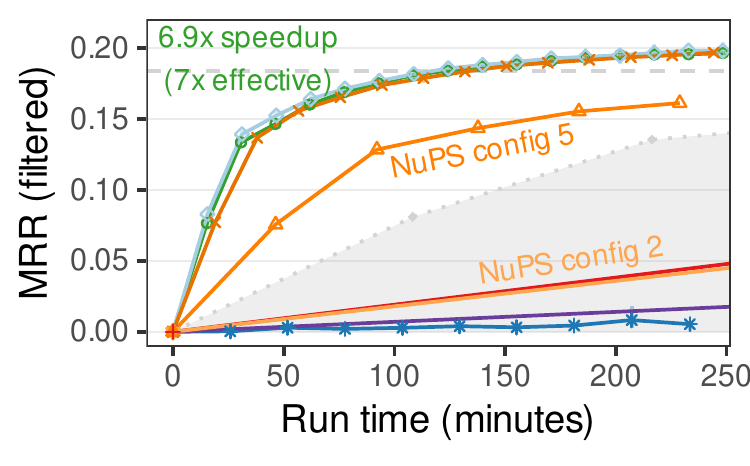}
    \caption{Knowledge graph embeddings (KGE)}
    \label{fig:ete:kge}
  \end{subfigure}%
  \begin{subfigure}[b]{0.33\textwidth}
    \centering
    \includegraphics[page=1,width=0.95\textwidth]{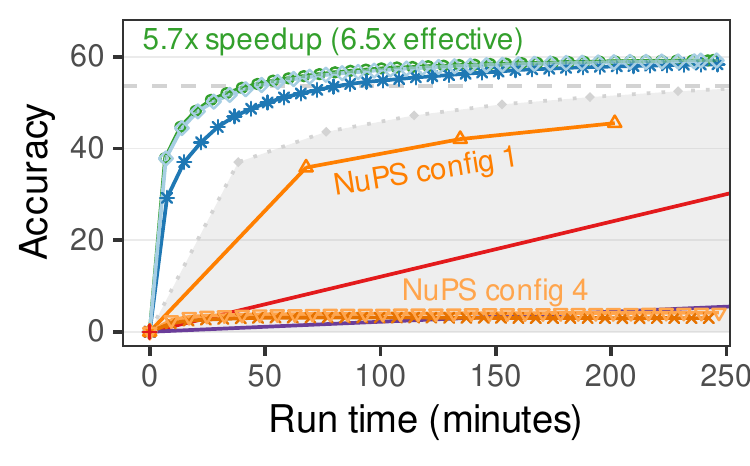}
    \caption{Word vectors (WV)}
    \label{fig:ete:wv}
  \end{subfigure}%
  \begin{subfigure}[b]{0.33\textwidth}
    \centering
    \includegraphics[page=1,width=0.95\textwidth]{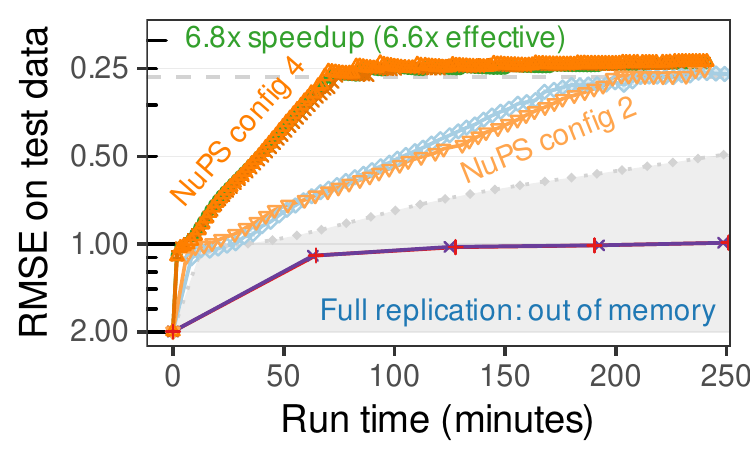}
    \caption{Matrix factorization (MF)}
    \label{fig:ete:mf}
  \end{subfigure}%

  \begin{subfigure}[b]{0.28\textwidth}
    \centering

    \def \lwidth {0.88}
    \def \lspace {\vspace{0.09cm}}

    \includegraphics[page=1,width=\lwidth\textwidth, clip, trim=0cm 3.6cm 0cm 0cm]{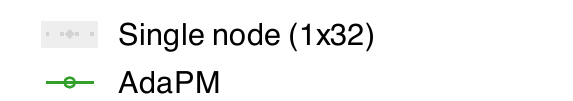} 

    \lspace

    \includegraphics[page=1,width=\lwidth\textwidth, clip, trim=0cm 0cm 0cm 3.1cm]{./plots/ete_legend_grid_all.pdf} 

    \lspace

    \includegraphics[page=1,width=\lwidth\textwidth, clip, trim=0cm 2.6cm 0cm 1.02cm]{./plots/ete_legend_grid_all.pdf} 

    \lspace

    \includegraphics[page=1,width=\lwidth\textwidth, clip, trim=0cm 1.54cm 0cm 2.06cm]{./plots/ete_legend_grid_all.pdf} 

    \vspace{0.2cm}
  \end{subfigure}%
  \begin{subfigure}[b]{0.33\textwidth}
    \centering
    \includegraphics[page=1,width=0.95\textwidth]{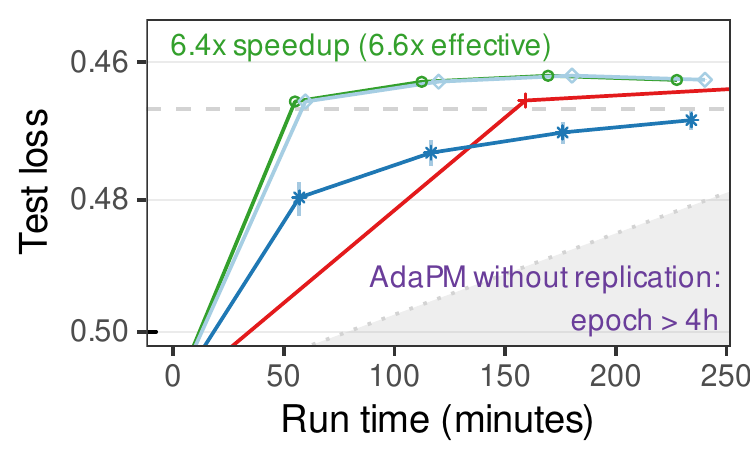}
    \caption{Click-through-rate predication (CTR)}
    \label{fig:ete:ctr}
  \end{subfigure}%
  \begin{subfigure}[b]{0.33\textwidth}
    \centering
    \includegraphics[page=1,width=0.95\textwidth]{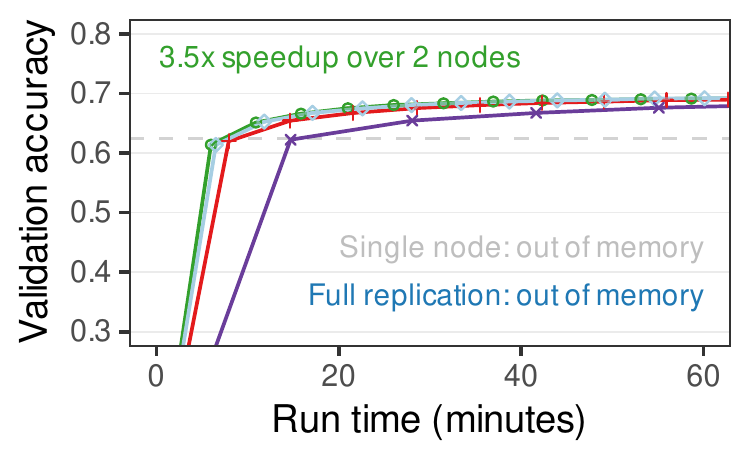}
    \caption{Graph neural networks (GNN)}
    \label{fig:ete:gnn}
  \end{subfigure}%
  \begin{subfigure}[b]{0.06\textwidth}
    \ 
  \end{subfigure}%

  \caption{Performance of \sys{} on 8 nodes (32 threads per node), compared to
    efficient single-node baselines (Section~\ref{sec:exp:single}), manually
    tuned PM (Section~\ref{sec:exp:nups}), standard PM approaches
    (Section~\ref{sec:exp:standard}), and single-technique \sys{} variants
    (Section~\ref{sec:exp:ablation}).
  }
  \label{fig:ete}
  \figurespace{}
\end{figure*}


\subsection{Overall Performance (Figure~\ref{fig:ete})}
\label{sec:exp:single}
We ran \sys{} on all tasks, without any tuning, and compared its performance to
the one of a single node. Figure~\ref{fig:ete} depicts the
results. 
\iflong
Figure~\ref{fig:ete:epoch} (page~\pageref{fig:ete:epoch} of the appendix) 
\fi
additionally shows model quality per epoch. 
\textbf{\sys{} achieved good speedups over the single node
	for all tasks out of the box, with 6.5x--7.0x effective speedups on 8 nodes.}

We measured these speedups against the efficient shared-memory single-node
implementation to ensure that they are practically relevant. Not comparing the
performance of distributed implementations to single-node baselines or comparing
to inefficient single-node implementations can be misleading \cite{lapse}.

\subsection{Comparison to Manually Tuned PM (Figure~\ref{fig:ete})}
\label{sec:exp:nups}
We further compared the performance of \sys{} to NuPS on the tasks for which NuPS
implementations and tuned configurations are available (KGE, WV, and MF).
Figures~\ref{fig:ete:kge}, \ref{fig:ete:wv}, and \ref{fig:ete:mf} depict the
results. 
\textbf{\sys{} matched or even outperformed the performance of NuPS
  across all tasks.}

NuPS required task-specific tuning to achieve good performance. The figures
depict three of the six NuPS configurations: (i) the best and worst
performing ones \emph{per task} from our
hyperparameter search and (ii) the ones
tuned by the NuPS authors. Different configurations were efficient
for different tasks. For example, configuration~4 was the best one
for MF, but the worst one for WV.

\sys{} matched (MF) or outperformed (slightly in KGE\footnote{Epoch were 20\%
  faster in \sys{} than in tuned NuPS.} and drastically in WV)
the best NuPS configurations. \sys{} can outperform NuPS because \sys{} manages
parameters more precisely, e.g., it maintained replicas only
while needed, allowing it to synchronize fewer replicas more frequently.
\iflong
Appendix~\ref{sec:exp:in-action}
\fi
provides more insight into how \sys{} works and how it differs from NuPS.

\subsection{Comparison to Standard PM (Figure~\ref{fig:ete})}
\label{sec:exp:standard}
We compared \sys{} to static full replication and static parameter partitioning
for all tasks. Again, Figure~\ref{fig:ete} 
shows the results. \textbf{\sys{} outperformed standard PM.}

\textbf{Static full replication} provided poor performance for all tasks but
WV. It provided poor model quality for KGE and CTR because synchronizing full
replicas on all nodes allowed for only infrequent replica synchronization.
It ran out of memory for MF and GNN because their models are large%
.
It worked well for WV because the WV model is small (only \SI{14}{GB},  see 
\iflong
Table~\ref{tab:datasets}%
\fi
) and the WV task is robust towards
infrequent replica synchronization.

\textbf{Static parameter partitioning} was inefficient because it required
synchronous network communication for the majority of parameter accesses. It was
relatively efficient for GNN because this task accesses parameters in large
groups, such that the impact of access latency is small.

\begin{table}
  \caption{Per-epoch network communication and staleness.}
  \label{tab:comm-and-freshness}
  \centering
  \resizebox{\columnwidth}!{
  \begin{threeparttable}
    \setlength\tabcolsep{2.5pt}
    \begin{tabular}{lrrrrrrrrrr} \toprule

      Variant  & \multicolumn{5}{c}{Commun.~(GB per node)} & \multicolumn{5}{c}{Mean replica staleness (ms)} \\
      \cmidrule(lr){2-6} \cmidrule(lr){7-11}
               & KGE & WV & MF & CTR & GNN & KGE & WV & MF & CTR & GNN \\
      \midrule
                            \sys{}  & 490 & 415  & 39 & 910  & 82  & 1.2 & 11.2 & 1.4 & 7.2 & 2.0 \\
             \sys{} w/o relocation  & 685 & 607 & 349 & 1066 & 171 & 1.4 & 13.8 & 7.7 & 7.7 & 2.6 \\
      \bottomrule
    \end{tabular}
  \end{threeparttable}}
\end{table}


\subsection{Comparison to Single-Technique Adaptive PM (Figure~\ref{fig:ete})}
\label{sec:exp:ablation}

We compared \sys{} to two ablation variants, see Figure~\ref{fig:ete}. These variants are identical to
\sys{}, but each variant is restricted to one management technique:
\emph{\sys{} without relocation} only replicates parameters, and \emph{\sys{} without
  replication} only relocates parameters. \textbf{\sys{} without replication was \emph{inefficient};
  \sys{} without relocation was \emph{efficient} for most tasks.}

\textbf{\sys{} without replication} performed poorly for all tasks
because relocation is inefficient for hot spot parameters, as observed
previously~\cite{nups}.

\textbf{\sys{} without relocation} was efficient for KGE, WV, CTR, and GNN. For
MF, it was 3.0x 
slower than \sys{} because the MF task exhibits locality (due to
row-partitioning, each row parameter is accessed by only one node) and
replication is inefficient for managing locality.

\subsection{The Benefit of Relocation (Table~\ref{tab:comm-and-freshness})}
\label{sec:exp:why-relocation}

As replication-only \sys{} performed well on many tasks (see
Section~\ref{sec:exp:ablation}), we further investigated whether it is
beneficial for \sys{} to employ relocation in addition to replication. To do so,
we measured the amount of communicated data and replica
staleness. 
\add{For simplicity, we take the time since the last replica refresh (i.e., the last check for updates) as staleness and ignore whether or not the value has actually changed since then.}
Table~\ref{tab:comm-and-freshness} depicts the results. Besides
improving performance for some tasks (see Section~\ref{sec:exp:ablation}),
\textbf{employing relocation reduced network overhead and decreased
  replica staleness for all tasks.}

The contrast in communication and staleness was particularly large for
tasks with locality (MF and GNN due to data partitioning), where \sys{}
communicated up to 9x fewer data. But supporting
relocation also improved efficiency for all other tasks.
For example, \sys{} without relocation communicated 40\% more data for one KGE
epoch because relocation is more efficient if two nodes access a
parameter after each other: relocation sends the parameter directly from the
first to the second node, whereas replication synchronizes via the owner
node.

\begin{figure}
  \centering
  \panelfiguresetup{}
  \includegraphics[page=1,width=1.0\columnwidth]{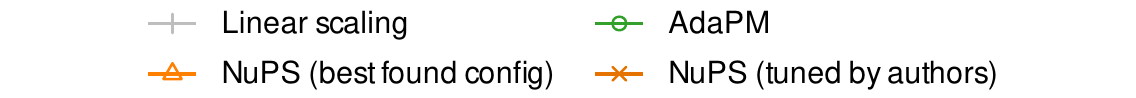}

  \begin{subfigure}[b]{.35\columnwidth}
    \centering
    \includegraphics[page=1,width=1.0\textwidth]{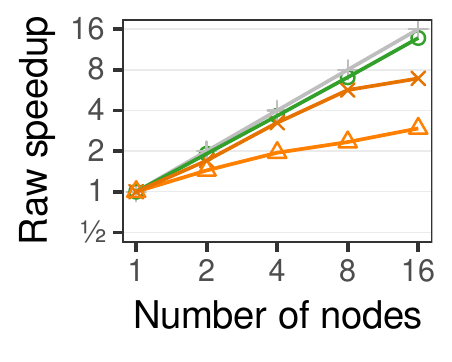}
    \caption{KGE: raw}
    \label{fig:scale:kge}
  \end{subfigure}%
  \begin{subfigure}[b]{.324\columnwidth}
    \centering
    \includegraphics[page=1,width=1.0\textwidth]{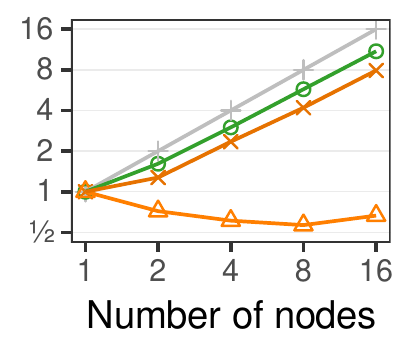}
    \caption{WV: raw}
    \label{fig:scale:wv}
  \end{subfigure}%
  \begin{subfigure}[b]{.324\columnwidth}
    \centering
    \includegraphics[page=1,width=1.0\textwidth]{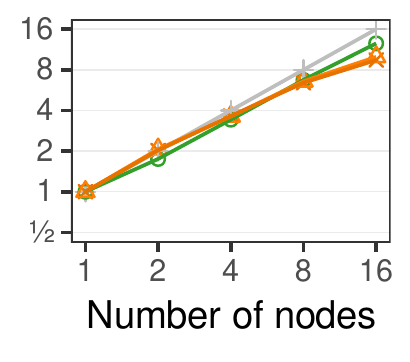}
    \caption{MF: raw}
    \label{fig:scale:mf}
  \end{subfigure}%

  \begin{subfigure}[b]{.35\columnwidth}
    \centering
    \includegraphics[page=1,width=1.0\textwidth]{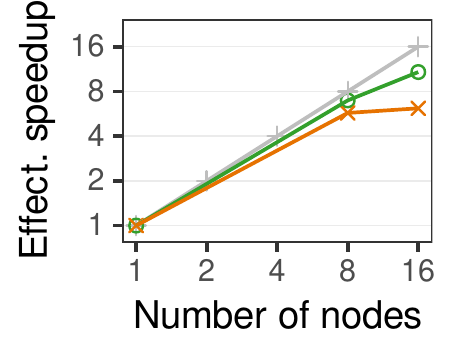}
    \caption{KGE: effective}
    \label{fig:eff-scale:kge}
  \end{subfigure}%
  \begin{subfigure}[b]{.324\columnwidth}
    \centering
    \includegraphics[page=1,width=1.0\textwidth]{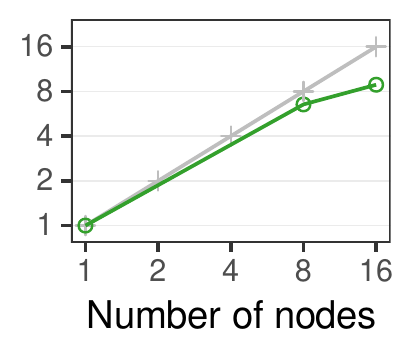}
    \caption{WV: effective}
    \label{fig:eff-scale:wv}
  \end{subfigure}%
  \begin{subfigure}[b]{.324\columnwidth}
    \centering
    \includegraphics[page=1,width=1.0\textwidth]{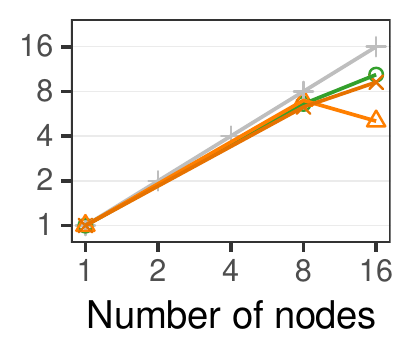}
    \caption{MF: effective}
    \label{fig:eff-scale:mf}
  \end{subfigure}%

  \caption{Scalability (logarithmic axes). Raw speedup (a-c), i.e., w.r.t.
    epoch run time, and effective speedup (d-f), i.e., w.r.t. reaching 90\% of
    the best model quality observed on a single node. Runs that did not
    reach this threshold within the time limit are not shown.}
  \label{fig:scale}
  \figurespace{}
\end{figure}

\subsection{Scalability (Figure~\ref{fig:scale})}
\label{sec:exp:scalability}

We investigated the scalability of \sys{} and compared it to NuPS.
Figure~\ref{fig:scale} depicts raw and effective speedups over the single node
for KGE, WV, and MF; 
\iflong
Figure~\ref{fig:scale:appendix} (page~\pageref{fig:scale:appendix}) 
\fi
does so for the other tasks.
    \textbf{\sys{} scaled
  efficiently, achieving near-linear raw and good effective speedups.}

\sys{} scaled more efficiently than NuPS because NuPS's scalability was limited
by relocation conflicts (i.e., concurrent accesses to a relocation-managed
parameter). For example, in KGE, the share of remote accesses in the best found
NuPS configuration was 1.2\%, 2.4\%, 3.4\%, and 5.3\% on 2, 4, 8, and 16 nodes,
respectively; in \sys{}, it was <0.0001\%. The effective speedups slightly
dropped on 16 nodes because we tuned task hyperparameters (e.g., learning rate
and regularization) for the single node and---to minimize the impact of
hyperparameter tuning---used these settings throughout all experiments.
These settings were not optimal for runs with 16x more parallelism. Other settings achieved
better effective scalability.

\begin{figure}
  \panelfiguresetup{}
  \centering

  \begin{subfigure}[b]{0.38\columnwidth}
    \centering
    \includegraphics[page=1,width=1.0\textwidth]{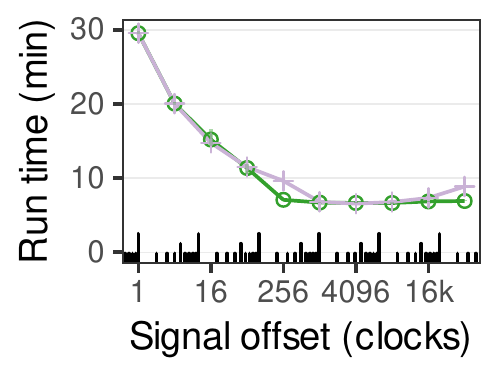}
    \caption{Epoch run time}
  \end{subfigure}%
  \begin{subfigure}[b]{0.38\columnwidth}
    \centering
    \includegraphics[page=1,width=1.0\textwidth]{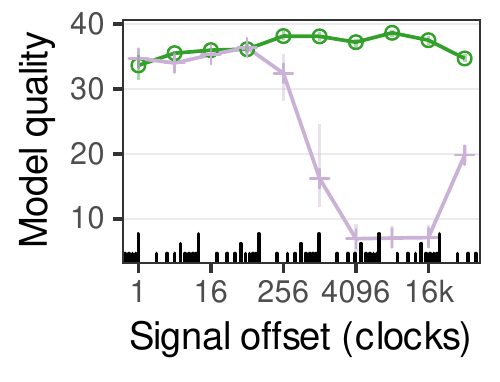}
    \caption{Model quality}
  \end{subfigure}%
  \begin{subfigure}[b]{0.24\columnwidth}
  \includegraphics[page=1,width=1.0\columnwidth]{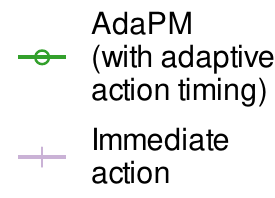}
  \vspace{1cm}
  \end{subfigure}%
  \caption{The effect of adaptive action timing for WV on epoch run time (a) and
    on model quality after one epoch (b).}
  \label{fig:action-timing:selected}
  \figurespace{}
\end{figure}

\subsection{Effect of Action Timing (Figure~\ref{fig:action-timing:selected})}
\label{sec:exp:action-timing}

To investigate the effect of action timing, we compared \sys{} to an ablation
variant that acts immediately after each intent signal, on workloads with varying
signal offsets. Figure~\ref{fig:action-timing:selected} depicts the results for
WV, see
\iflong
Figure~\ref{fig:action-timing} (page~\pageref{fig:action-timing})
\fi
for further tasks. 
\textbf{With adaptive action timing, \sys{} was efficient for any
  sufficiently large signal offset.}

\textbf{Early signals. } With adaptive action timing, \sys{} provided excellent
performance for all large signal offset values. In contrast, with immediate
action, performance was poor for large signal offsets: run time increased and
model quality decreased. The reason for this was that the
immediate action variant maintained replicas for longer than necessary (thus
lowering synchronization frequency).

\textbf{Late signals. } Smaller relocation offsets improved performance for
immediate action, but did not further improve performance for \sys{}. For
both, epoch run time was poor when intent was signaled so late that the
system did not have sufficient time for setting up replicas or relocating
parameters. Thus, there was a task-specific optimum value
for immediate action (necessitating tuning), but not for \sys{}.

\subsection{Comparison to GPU Implementations}
\label{sec:exp:task-specific}
CTR and GNN models are typically trained on GPUs, whereas we used CPUs in our
proof-of-concept implementation of \sys{}. To put our results in perspective, we
briefly compared to recent GPU-based methods for these two tasks. Some overhead
is expected because these methods are task-specific, i.e., they use
specialized and tuned training algorithms that cannot be used for other tasks directly,
whereas \sys{} is general-purpose. 
Although the CPU and GPU settings are very different and cannot be compared
directly, we found that \textbf{the results obtained by \sys{} seem
  competitive}.

For CTR, Zheng et al.~\cite{cowclip} ran one epoch on a 90\% train split of our
dataset, for the same model, and with comparable batch size\footnote{Zheng et
  al.~\cite{cowclip} used a batch size of \num{1000} with synchronous parallel
  SGD. For \sys{}, we used a batch size of \num{128} with 8 workers of
  asynchronous SGD.} in 76.8 minutes on one V100 GPU. \sys{} took 57.5 minutes on
8 CPU nodes for a 85.7\% train split. \add{The key bottleneck of GPU-based
  methods is data transfer to and from GPU memory as the models do not fit in
  GPU memory. Very large batch sizes can mitigate this to some extent, but
  require careful work to avoid loss of accuracy~\cite{cowclip}.}

For GNN, Min et al.~\cite{gcn-gpu-comm-arch} ran one epoch on the same
dataset, but with a different model, on 2 RTX 3090 GPUs in 10.2 minutes.
\sys{} ran 1 epoch in 5.3 minutes on 8 CPU nodes. MariusGNN~\cite{mariusgnn}
achieved a validation accuracy of 0.6638 in 8.2 minutes on the same dataset
(with a slightly different model) using 1 V100 GPU. Waleffe et
al.~\cite{mariusgnn} further report that DGL~\cite{dgl} required 4 V100
GPUs to achieve the same accuracy in similar time. \sys{} achieved slightly
better accuracy (0.665) after 15.5 minutes using 8 CPU nodes.


\section{Conclusion}

\add{We proposed intent signaling, a novel mechanism that decouples providing
  information about parameter accesses (simple) from how and when it is
  exploited (hard). Intent signaling increases ease of use and efficiency in
  parameter management.} We presented \sys{}, a parameter manager that
automatically adapts to the underlying ML task based solely on intent signals.
In our experimental study, \sys{} was efficient for many ML tasks out of the
box, without requiring any tuning, and matched or even outperformed
state-of-the-art (more complex to use) systems. Interesting
directions for future work are how to better integrate co-processor (e.g, GPU)
memory in an adaptive parameter manager such as \sys{} and how to support intent
signaling directly in common ML systems.


\section*{Acknowledgements}
  This work was supported by the German Ministry of Education and Research in
  the BIFOLD (01IS18037A) program and by the
  German Research Foundation in the moreEVS (410830482) program.

\bibliographystyle{ACM-Reference-Format}
\bibliography{main}

\iflong
\appendix

\def \iall {17}
\def \ihalf {8}
\def \ihalfpp {9}

\def \yoffset {6.5}
\def \xoffset {4.5}

\def \textsep {1.1}

\begin{figure*}
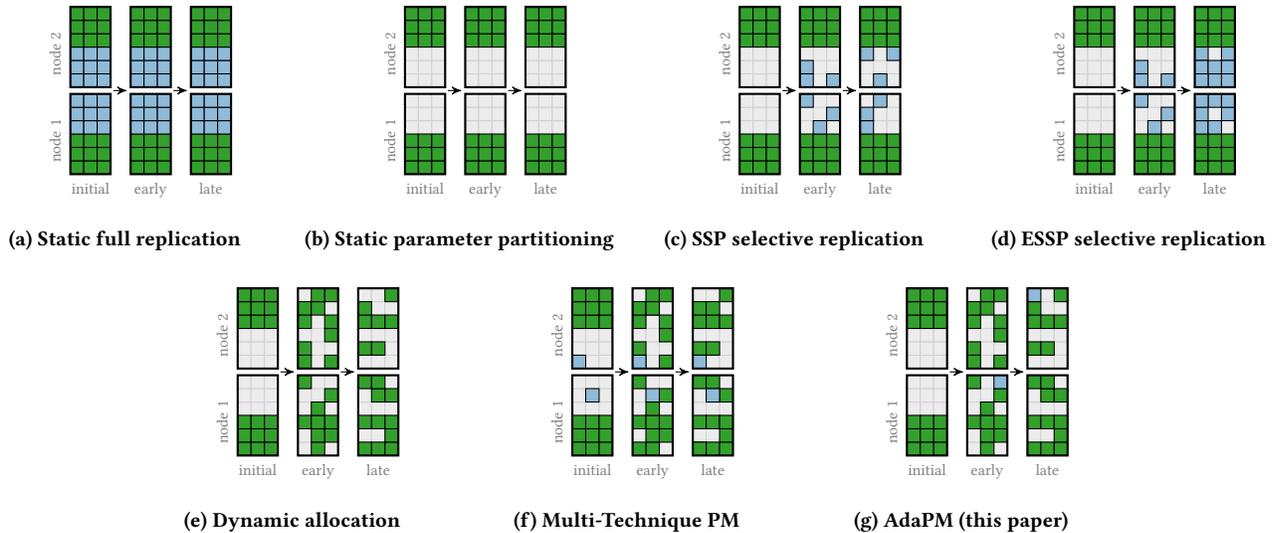

  \centering

  \captionsetup[subfigure]{aboveskip=0pt,belowskip=12pt} 
  
  \begin{subfigure}[c]{.25\textwidth}
    \centering

    \approachStart{}

    \foreach \i in {0,...,\ihalf}
    \repPstatic{\i}{1}
    \allocPstatic{\i}{0}
    ;

    \foreach \i in {\ihalfpp,...,\iall}
    \repPstatic{\i}{0}
    \allocPstatic{\i}{1}
    ;

    \approachEnd{}
    
    \caption{Static full replication}
    \label{fig:pm:full-replication}
  \end{subfigure}%
  \begin{subfigure}[c]{.25\textwidth}
    \centering

    \approachStart{}

    \foreach \i in {0,...,\ihalf}
      \allocPall{\i}{0};
    \foreach \i in {\ihalfpp,...,\iall}
      \allocPall{\i}{1};

    \approachEnd{}
    
    \caption{Static parameter partitioning}
    \label{fig:pm:classic}
  \end{subfigure}%
  \begin{subfigure}[c]{.25\textwidth}
    \centering

    \approachStart{}

    \foreach \i in {0,...,\ihalf}
    \allocPall{\i}{0};
    \foreach \i in {\ihalfpp,...,\iall}
    \allocPall{\i}{1};

    \repPsingle{3}{1}{1};
    \repPsingle{2}{1}{1};
    \repPsingle{0}{1}{1};
    \repPsingle{15}{0}{1};
    \repPsingle{14}{0}{1};
    \repPsingle{10}{0}{1};

    \repPsingle{1}{1}{2};
    \repPsingle{6}{1}{2};
    \repPsingle{8}{1}{2};
    \repPsingle{9}{0}{2};
    \repPsingle{12}{0}{2};
    \repPsingle{16}{0}{2};

    \approachEnd{}

    \caption{SSP selective replication}
    \label{fig:pm:ssp}
  \end{subfigure}%
  \begin{subfigure}[c]{.25\textwidth}
    \centering

    \approachStart{}

    \foreach \i in {0,...,\ihalf}
    \allocPall{\i}{0};
    \foreach \i in {\ihalfpp,...,\iall}
    \allocPall{\i}{1};

    \repPsingle{3}{1}{1};
    \repPsingle{2}{1}{1};
    \repPsingle{0}{1}{1};
    \repPsingle{15}{0}{1};
    \repPsingle{14}{0}{1};
    \repPsingle{10}{0}{1};

    \repPsingle{1}{1}{2};
    \repPsingle{6}{1}{2};
    \repPsingle{8}{1}{2};
    \repPsingle{9}{0}{2};
    \repPsingle{12}{0}{2};
    \repPsingle{16}{0}{2};

    \repPsingle{0}{1}{2};
    \repPsingle{3}{1}{2};
    \repPsingle{2}{1}{2};
    \repPsingle{15}{0}{2};
    \repPsingle{14}{0}{2};
    \repPsingle{10}{0}{2};

    \repPsingle{17}{0}{2};
    \repPsingle{4}{1}{2};
    \repPsingle{5}{1}{2};

    \approachEnd{}

    \caption{ESSP selective replication}
    \label{fig:pm:essp}
  \end{subfigure}%

  \begin{subfigure}[c]{.25\textwidth}
    \centering
    \pgfmathsetseed{26}

    \approachStart{}

    \foreach \i in {0,...,\ihalf}
    \allocP{\i}{0}{0};
    \foreach \i in {\ihalfpp,...,\iall}
    \allocP{\i}{1}{0};

    \allocP{0}{1}{1}; 
    \allocP{1}{0}{1};
    \allocP{2}{1}{1}; 
    \allocP{3}{1}{1}; 
    \foreach \i in {4,...,\ihalf}
      \allocP{\i}{0}{1};
    \allocP{8}{1}{1};
    \allocP{9}{1}{1};
    \allocP{10}{0}{1}; 
    \allocP{11}{1}{1};
    \allocP{12}{1}{1};
    \allocP{13}{1}{1};
    \allocP{14}{0}{1}; 
    \allocP{15}{0}{1}; 
    \allocP{16}{1}{1};
    \allocP{17}{1}{1};

    \foreach \i in {0,...,\iall} {
      \pgfmathsetmacro{\node}{random(0,1)}
      \pgfmathsetmacro{\node}{random(0,1)}
      \allocP{\i}{\node}{2};
    }

    \approachEnd{}

    \caption{Dynamic allocation}
    \label{fig:pm:lapse}
  \end{subfigure}%
  \begin{subfigure}[c]{.25\textwidth}
    \centering
    \pgfmathsetseed{26}

    \approachStart{}

    \foreach \i in {0,...,\ihalf}
    \allocP{\i}{0}{0};
    \foreach \i in {\ihalfpp,...,\iall}
    \allocP{\i}{1}{0};

    \foreach \i in {0,...,\iall} {
      \pgfmathsetmacro{\node}{random(0,1)}
      \pgfmathsetmacro{\node}{random(0,1)}
      \allocP{\i}{\node}{2};
    }

    \allocP{0}{1}{1}; 
    \allocP{1}{0}{1};
    \allocP{2}{1}{1}; 
    \allocP{3}{1}{1}; 
    \foreach \i in {4,...,\ihalf}
    \allocP{\i}{0}{1};
    \allocP{8}{1}{1};
    \allocP{9}{1}{1};
    \allocP{10}{0}{1}; 
    \allocP{11}{1}{1};
    \allocP{12}{1}{1};
    \allocP{13}{1}{1};
    \allocP{14}{0}{1}; 
    \allocP{15}{0}{1}; 
    \allocP{16}{1}{1};
    \allocP{17}{1}{1};

    \repPstatic{13}{0};
    \allocPstatic{13}{1};

    \repPstatic{0}{1};
    \allocPstatic{0}{0};

    \approachEnd{}

    \caption{Multi-Technique PM}
    \label{fig:pm:nups}
  \end{subfigure}%
  \begin{subfigure}[c]{.25\textwidth}
    \centering
    \pgfmathsetseed{26}

    \approachStart{}

    \foreach \i in {0,...,\ihalf}
    \allocP{\i}{0}{0};
    \foreach \i in {\ihalfpp,...,\iall}
    \allocP{\i}{1}{0};

    \foreach \i in {0,...,\iall} {
      \pgfmathsetmacro{\node}{random(0,1)}
      \pgfmathsetmacro{\node}{random(0,1)}
      \allocP{\i}{\node}{2};
    }

    \allocP{0}{1}{1}; 
    \allocP{1}{0}{1};
    \allocP{2}{1}{1}; 
    \allocP{3}{1}{1}; 
    \foreach \i in {4,...,\ihalf}
    \allocP{\i}{0}{1};
    \allocP{8}{1}{1};
    \allocP{9}{1}{1};
    \allocP{10}{0}{1}; 
    \allocP{11}{1}{1};
    \allocP{12}{1}{1};
    \allocP{13}{1}{1};
    \allocP{14}{0}{1}; 
    \allocP{15}{0}{1}; 
    \allocP{16}{1}{1};
    \allocP{17}{1}{1};

    \repPsingle{17}{0}{1};
    \repPsingle{15}{1}{2};

    \approachEnd{}

    \caption{\sys{} (this paper)}
    \label{fig:pm:adaps}
  \end{subfigure}%

  \newcommand*\singlebase[0]{\tikz[baseline=0.3pt,
    scale=0.2]{
      \baseP{0}{0}{0};
    }}
  \newcommand*\singlealloc[0]{\tikz[baseline=0.3pt,
    scale=0.2]{
      \allocP{0}{0}{0};
    }}
  \newcommand*\singlerep[0]{\tikz[baseline=0.3pt,
    scale=0.2]{
      \repPsingle{0}{0}{0};
    }}

  \caption{Parameters held by different nodes at different times (initially;
    and early and late during training) in common parameter management approaches.
    One square depicts one parameter. A node either cannot access the
    parameter locally (\protect\singlebase{}), holds the main copy of the
    parameter (\protect\singlealloc{}), or holds a replica of the
    parameter (\protect\singlerep{}).}
  \label{fig:existing-approaches}
  \figurespace{}
\end{figure*}


\section{Efficiency and Complexity of Existing Approaches}
\label{sec:parameter-management-approaches-long}

In the following, we analyze existing general-purpose PM approaches w.r.t. their
ease of use, their efficiency for sparse ML tasks (i.e., tasks in which each
update step accesses only a small subset of all model parameters), and in which aspects they
are adaptive. Figure~\ref{fig:existing-approaches} illustrates existing
approaches. Table~\ref{tab:approaches-to-parameter-management}
(page~\pageref{tab:approaches-to-parameter-management}) summarizes our analyses.

\subsection{Static Full Replication}
Static full replication replicates the full model to all cluster nodes
statically (i.e., throughout training) and synchronizes the replicas
periodically, either synchronously (triggered by the application) or
asynchronously in the background. This approach is currently popular for
training dense neural networks~\cite{cnn, horovod,
  pytorch-distributed}, in particular with synchronous synchronization after
stochastic gradient descent (SGD) mini-batches. As it is entirely static, it
requires no run time information from the application. So it is relatively easy
to use. (It does, however, require the application to either trigger replica
synchronization or to set the frequency of background synchronization.)
Parameter access is fast, as every worker can access every parameter locally,
without synchronous network communication. However, full replication is
communication-inefficient for sparse workloads~\cite{nups}, as it maintains the
replicas of all parameters on all nodes throughout the training task, even
though each node accesses only a small subset of these replicas at each point in
time. Also, full replication limits model size to the memory capacity of a
single node. \textbf{In summary, full replication is very easy to use, but
  inefficient for sparse workloads because it over-communicates.}

\subsection{Static Parameter Partitioning}
Static parameter partitioning (as done in classic parameter servers~\cite{smola10, ahmed12, pslite}) partitions the model
parameters to the cluster nodes and provides global reads and writes to the
parameters by transparently sending messages to the corresponding nodes. No
replicas are created. Static parameter partitioning is easy to use: it requires no
information from the application and no hyperparameter tuning. However, the
approach is inefficient because the vast majority of parameter accesses involve
synchronous network communication for sending messages to the node that holds
the parameter~\cite{lapse}. \textbf{In summary, static parameter partitioning is
  very easy to use, but inefficient due to synchronous network communication.}

\subsection{Selective Replication}
A selective replication parameter manager, such as Petuum~\cite{petuum},
statically partitions parameters. During training, it
adaptively replicates a subset of the parameters to nodes that access these
parameters. Potentially, this makes PM more efficient, as replicas can be used
to process repeated accesses to the same parameter locally. Petuum sets up a
replica for a specific parameter at a specific node reactively when a worker on
that node accesses the parameter. Thus, the workers have to wait for replicas to
be set up synchronously.

Petuum supports two heuristics to decide how long to maintain a replica at a
node after parameter access: SSP~\cite{ssp} and ESSP~\cite{essp}. SSP maintains
a replica for an application-specified number of logical clocks, the so-called
\emph{staleness bound}. Applications have to tune this staleness bound
specifically for each task, as the staleness bound impacts both model quality
and run time efficiency. This tuning makes SSP complex to use. Further, SSP is
inefficient for many tasks because, for realistic staleness bounds, no replicas
are set up for the majority of parameter accesses, so that workers have to wait
for synchronous replica setup frequently. \textbf{In summary, an SSP parameter
  manager is complex to use because it requires tuning and inefficient because
  of synchronous replica setup.}

ESSP is a coarse heuristic that also creates a replica when a parameter is
accessed at a node, and then maintains the replica \emph{throughout the entire
  training task}. This mitigates the inefficiency of reactive replica creation
(as each replica is set up only once), at the cost of over-communication. For
many workloads, after a short setup phase, ESSP is essentially equivalent to
full replication, i.e., it holds a full model replica on every node, although
only a small part of the model is accessed at every point in time. This makes
ESSP inefficient for sparse workloads~\cite{nups}. \textbf{In summary, an ESSP
  parameter manager is complex to use, and inefficient due to
  over-communication.}

\subsection{Dynamic Parameter Allocation}
A dynamic parameter allocation parameter manager, such as Lapse~\cite{lapse},
adaptively changes the location of parameters during training, i.e., it
relocates parameters among nodes. The advantage of relocating is that parameters
can then be accessed locally, without network communication, at the respective
nodes, and no replica synchronization is required. Key for the efficiency of a
dynamic allocation PM is that parameter relocation is \emph{proactive}, i.e.,
that parameter relocation runs asynchronously and is finished before the
parameter is accessed. PM lacks the necessary information to initiate
relocations proactively. Thus, Lapse requires the application to initiate
parameter relocations manually via an additional \texttt{localize} primitive.
Application developers are required to add appropriate invocations to their
application code and---for optimal performance---tune the \emph{relocation
  offset}, i.e., how long before the actual access parameter relocation is
initiated. Relocation should be early enough such that it is finished when the
parameter is accessed, but not too early to minimize the probability of
relocating the parameter away from other nodes that are still accessing it.
Offloading these performance-critical decisions to the application makes dynamic
allocation PM complex to use and leads to potentially sub-optimal performance.
In addition, dynamic allocation PM is inefficient for many real-world ML tasks
because they are inefficient for hot spots, i.e., parameters that are frequently
accessed by multiple nodes concurrently~\cite{nups}. \textbf{In summary, dynamic
  parameter allocation is complex to use, and inefficient for many ML tasks.}

\subsection{Multi-Technique Parameter Management}

Multi-technique parameter managers support multiple parameter management techniques (e.g.,
replication, static partitioning, or relocation) and let the application pick a suitable
one for each parameter. Parallax~\cite{parallax} and BiPS~\cite{bips} support
static full replication (i.e., creating replicas on all nodes) and static partitioning.
NuPS~\cite{nups} supports static full
replication and dynamic parameter allocation. The choice of technique
is static: for each parameter, the application picks one technique before
training, which is then used throughout. 
Using a suitable technique for each parameter can improve PM
efficiency~\cite{nups,bips,parallax}. However, it requires information about the
workload. As these parameter managers decide on a technique for each parameter before training,
they require this information upfront. There are heuristics that pick a
technique for each parameter~\cite{parallax, nups}. These heuristics require
access frequency statistics and do not consistently achieve optimal
performance. Thus, manual tuning can be required to
achieve high efficiency. These information and tuning requirements make multi-technique PM very
complex to use and---if not tuned appropriately---lead to
sub-optimal performance. NuPS additionally requires the application to manually
trigger relocations (as in Lapse), further complicating its use.
\textbf{In summary, multi-technique PM can be efficient, but is very complex to use.}


\section{\sys{} Architecture Details}

\subsection{Responsibility Follows Allocation}
\label{sec:sys:owner-node}

Based on intent signals, \sys{} decides when to relocate a parameter and when to
maintain a replica on which node. Two important design decisions to enable this
precise management are: (i) which node makes these decisions and
(ii) when replicas exist, how to keep them synchronized (efficiently)? A key
feature of \sys{} is that the node at which a parameter is currently
allocated---the \emph{owner node}\footnote{We adopt this term from
  Lapse.} of the parameter---takes the main responsibility for
both. The owner node decides whether to relocate a parameter and where to
maintain replicas (Section~\ref{sec:sys:decision-making-node}); and the owner
node acts as a hub for replica synchronization
(Section~\ref{sec:sys:replica-sync}). Placing responsibility at the owner node
reduces network overhead and can reduce processing load because the owner node
changes whenever the parameter is relocated, such that responsibility is close
to where the parameter and the associated processing is.

\subsubsection{Choice of Management Technique}
\label{sec:sys:decision-making-node}

\begin{figure}[t]
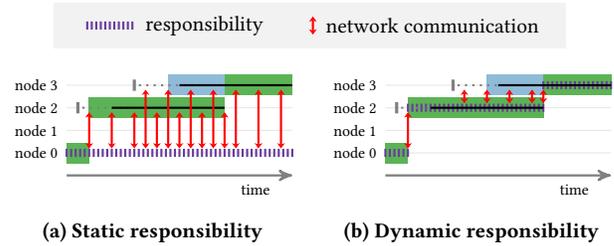


  \center
  \policyplotbase{Worker2}
  \draw [fill=gray!10,draw=none] (-1.5,-0.5) rectangle (21.2, 1.5);
  \responsibility{0}{0}{2}
  \node [anchor=west] at (2.2,0.5) {\small responsibility};
  \comm{-0.6}{0.6}{10}{10.5}
  \node [anchor=west] at (10.2,0.5) {\small network communication};

  \node [anchor=south] at (10,-0.8) {};
  \policyplotend{}

  \vspace{0.2cm}
  
  \begin{subfigure}[c]{.5\columnwidth}
    \centering
    \adapolicyhalf{Worker2}
    \alloc{0}{0}{1}
    \alloc{2}{1}{6}
    \alloc{3}{7}{3}
    \preprep{3}{5}{2}

    \responsibility{0}{0}{10}

    \signal{2}{0.5}{2.5}
    \intent{2}{2}{5}

    \signal{3}{3}{3}
    \intent{3}{5}{5}

    \comm{0}{2}{1}{7}
    \comm{0}{3}{3.5}{9.5}

    \policyplotend{}
    \caption{Static responsibility}
    \label{fig:commreq:static}
  \end{subfigure}%
  \begin{subfigure}[c]{.5\columnwidth}
    \centering
    \adapolicyhalf{Worker2}
    \alloc{0}{0}{1}
    \alloc{2}{1}{6}
    \alloc{3}{7}{3}
    \preprep{3}{5}{2}

    \responsibility{0}{0}{1}
    \responsibility{2}{1}{6}
    \responsibility{3}{7}{3}

    \signal{2}{0.5}{2.5}
    \intent{2}{2}{5}

    \signal{3}{3}{3}
    \intent{3}{5}{5}

    \comm{0}{2}{1}{1}
    \comm{2}{3}{3.5}{6.5}
    \comm{2}{3}{7}{7}
    
    \policyplotend{}
    \caption{Dynamic responsibility}
    \label{fig:commreq:dynamic}
  \end{subfigure}%
  \caption{Network communication for placing management responsibility
    (a) on the statically assigned home node (node~0) or (b) on the
    dynamically changing owner node. }
  \label{fig:commreq}
  \figurespace{}
\end{figure}


\sys{} chooses a technique based on the intent signals
of all nodes for a parameter. Thus, the intent signals of all nodes for one
parameter need to come together at one node, such that this node can make this
decision.

This decision could be made by a node that is \emph{statically} assigned to a
parameter (e.g., by hash partitioning). We call such a statically assigned node
the \emph{home node} \cite{steen17}. In the static
approach, nodes continuously send their intent signals for parameter~$k$ to the
parameter's home node. The home node decides whether to relocate or replicate
the parameter, and instructs the current owner of parameter $k$ to act accordingly.
The advantage of this
static approach is that it is straightforward to route intent signals, as the
signals for one key are sent to the same node throughout training. Its
disadvantage is that the home node is always involved, even when it does not
have intent itself (a common disadvantage of home-based approaches in
distributed systems~\cite{steen17}). Figure~\ref{fig:commreq:static}
illustrates the communication for the static approach. Even while there is intent
only at node~2 and the parameter is allocated at node~2, node~2 needs to
communicate its intent to the home node (node~0), such that the home node can
decide to keep the parameter allocated at node~2. While there is intent at
nodes~2 and~3, both nodes need to communicate their intent to node~0.

To overcome this problem, \sys{} makes these decisions on the owner node of the
parameter, i.e., the node where the parameter is currently allocated. This node
changes whenever the parameter is relocated. The key advantage of this approach
is that the home node does not need to be involved, reducing network traffic and
processing load. Figure~\ref{fig:commreq:dynamic} illustrates the communication
for this dynamic approach. After an initial communication of intent (and a
subsequent relocation), no further
communication between node~2 and node~0 is necessary, as node~2 makes any
decisions about the parameter locally. While there is intent on node~2 and
node~3, node~3 communicates its intent directly to node~2, without involving node~0. A
disadvantage of the dynamic approach is that routing becomes more complex, as
the owner node changes throughout training. To overcome this disadvantage,
\sys{} employs location caches, which enables nodes to send their intent signals
to the current owner node directly, most of the time, see Section~\ref{sec:sys:routing}.

\subsubsection{Replica Synchronization}
\label{sec:sys:replica-sync}

While a parameter is replicated, multiple nodes hold a copy of the value of the
parameter and write updates to this local copy. For convergence, it is crucial
that these copies are synchronized, i.e., that updates of one node are
propagated to the replicas on other nodes. \sys{} employs relocation and
selective replication (as discussed in
Section~\ref{sec:sys:relocation-and-selective-replication}). Consequently, for
the majority of parameters, at a given time, only few nodes (if at all) hold a
replica for a given parameter. Thus, replica updates have to be propagated only
to a small subset of all nodes. Further, this subset is different for each
parameter and changes constantly and potentially rapidly. These two properties
make AllReduce or gossip-based synchronization approaches unattractive for
\sys{}. Instead, \sys{} propagates replica updates via the owner node of a
parameter: replica holders send updates to the parameter's owner, which then
propagates them to other replica holders.

To use the network efficiently, \sys{} batches replica updates (as, e.g.,
Petuum~\cite{petuum} does). As Petuum, \sys{} does
so for both directions of update propagation: from replica holders to the
owner node and from the owner node to the replica holders. To further improve
efficiency, \sys{} versions parameter values and communicates deltas: when nodes
request updates for their replicas, they include the version number
that is locally available; the owner node sends only those updates that the requesting
node has not received previously.

\subsection{Efficient Communication}
\label{sec:sys:efficient-communication}

\sys{} communicates to exchange intent signals, to relocate parameters, to
set up and destruct replicas, and to synchronize replicas. Key for the overall efficiency of \sys{} is that this communication is
efficient. In this section, we discuss several design aspects of \sys{} that
improve communication efficiency: \sys{} locally aggregates intent signals and
sends aggregated intent signals over the network
(Section~\ref{sec:how-to-communciate-intent}), groups messages
(Section~\ref{sec:sys:grouped-messages}), employs location caches for more
efficient routing (Section~\ref{sec:sys:routing}), and
relocates parameters only when there is exactly one node with active intent (Section~\ref{sec:sys:relocate-condition}).

\subsubsection{Aggregated Intent}
\label{sec:how-to-communciate-intent}

As discussed in Section~\ref{sec:sys:decision-making-node}, for each parameter,
there is one (dynamically changing) node that decides whether to relocate or
replicate a specific parameter. To enable
this decision, all other nodes need to continuously send their intent signals
for this specific parameter to this decision-making node.

A naive approach to intent communication is that all workers eagerly send each
intent (i.e., a tuple of parameter, start clock, end clock, intent
type\footnote{\sys{} does not need to communicate intent type, as it does not
  require to know the intent type for its decisions, see
  Section~\ref{sec:sys:relocation-and-selective-replication}.}, and worker
id\footnote{So that the system knows which worker's clock the start and end
  clocks refer to, and on which node the intent was signaled.}) to the node that
makes decisions immediately after the intent is
signaled. Additionally, workers regularly send the state of their clocks
(potentially by piggybacking on other messages). The decision-making node stores
all intents, determines which of them are active, and decides based on this perfect
view of all intents for this specific parameter. However, this naive approach
induces significant network overhead, as a large number of intent signals have
to be sent over the network constantly. Especially for hot spot parameters, for
which there are many intent signals, this can be prohibitive.

\sys{} employs a more communication-efficient approach: each node stores
inactive intents locally, determines which ones should be treated as active, and
sends aggregated information about active intents to the decision-making node.
More precisely, each node communicates to the decision-making node when
intent becomes active and it communicates when intent
expires. The node does not communicate which or how many workers have active
intent. This requires significantly less network communication, especially for
hot spot parameters. The disadvantage is that the decision-making node has less
information. Precisely, it does not
know about inactive intents or how long active intent will last. For the
decisions that \sys{} makes, this information is not required, such that \sys{}
adopts the more communication-efficient approach.

\subsubsection{Message Grouping}
\label{sec:sys:grouped-messages}

Adaptive action timing ensures that a parameter is relocated or a replica is
set up asynchronously, i.e., before a worker accesses the parameter. This allows
\sys{} to improve network efficiency by grouping messages for communicating
(aggregated) intent signals, for relocating parameters, and for creating,
destructing, and synchronizing replicas into
one request--response message protocol.

In more detail, to send a synchronization request, a dedicated thread at a node
collects (i) a list of parameters for which local workers have active intent and
(ii) all updates to local replicas. The node sends these to
the owners of the corresponding parameters in a \emph{synchronization request}.
Each owner (i) merges the replica updates into its parameter store and (ii)
responds to each intent signal (with parameter relocation or replica setup). It does so
in one (grouped) \emph{synchronization response}.
By default, \sys{} triggers a synchronization request as soon as the last
communication round
has finished. To reduce the network and CPU load for synchronization, \sys{}
allows for limiting the number of communication rounds per second.

\subsubsection{Routing}
\label{sec:sys:routing}

In \sys{}, nodes send intent signals and replica updates to the current owner node.
This owner node can change dynamically during run time. To route messages,
\sys{} adapts the home node forwarding approach (with location caches) of
Lapse~\cite{lapse}. We briefly describe this approach below and refer
to Renz-Wieland et al.~\cite{lapse} for a more detailed discussion.

As fallback, there is one \emph{home node} for each parameter. This home node
is assigned statically to each parameter (by hash partitioning). The home node
knows which (other) node is currently the owner of the parameter. If any node
does not know where a parameter is currently allocated, it sends its message to
the home node, which then forwards the message to the current owner. Whenever a
parameter is relocated, the old owner node informs the home node of this
relocation. These location updates are piggybacked onto synchronization
messages.

To increase efficiency, \sys{} additionally employs location caches. I.e., each
node locally stores the last known location for parameters that it accessed in
the past. This allows for sending updates and intent signals directly to the current
owner. \sys{} uses synchronization responses, outgoing parameter relocations,
and responses to remote parameter accesses to update location caches. It does
not explicitly invalidate location caches. Instead, it tolerates that messages can
be routed based on stale ownership information and relies on the receiving nodes to
forward the messages to the current owner (via the home node, see Renz-Wieland
et al.~\cite{lapse}
for a more detailed discussion). Location caches are more important in \sys{}
than in Lapse as there are scenarios in \sys{} in which nodes repeatedly send
messages to the owner node. In particular for hot spot parameters,
the owner node changes rarely (see, e.g., Figure~\ref{fig:scenario:hotspot}),
such that nodes send their signals and updates to the same owner node
repeatedly.

\subsubsection{Parameter Relocation}
\label{sec:sys:relocate-condition}

\begin{figure}
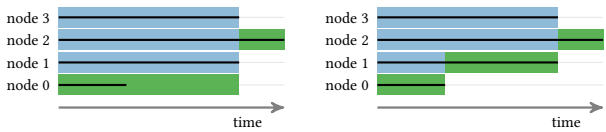

  \begin{subfigure}[c]{.5\columnwidth}
    \centering

    \adapolicyhalf{Worker2}

    \rep{1}{0}{8}
    \intent{1}{0}{8}

    \rep{2}{0}{8}
    \alloc{2}{8}{2}
    \intent{2}{0}{10}

    \rep{3}{0}{8}
    \intent{3}{0}{8}

    \alloc{0}{0}{8}
    \intent{0}{0}{3}

    \policyplotend{}

    \caption{Relocate only when exactly\,\,\,\, one node has active intent}
    \label{fig:owner-intent-stops:relocate-one}
  \end{subfigure}%
  \begin{subfigure}[c]{.5\columnwidth}
    \centering

    \adapolicyhalf{Worker2}

    \alloc{0}{0}{3}
    \intent{0}{0}{3}

    \rep{1}{0}{3}
    \alloc{1}{3}{5}
    \intent{1}{0}{8}

    \rep{2}{0}{8}
    \alloc{2}{8}{2}
    \intent{2}{0}{10}

    \rep{3}{0}{8}
    \intent{3}{0}{8}

    \policyplotend{}

    \caption{Relocate immediately when the owner's intent expires}
    \label{fig:owner-intent-stops:relocate-right-away}
  \end{subfigure}%
  \caption{\sys{} relocates a parameter only when there is exactly one node with
    active intent (and the parameter is currently not allocated at this node).}
  \label{fig:relocate-while-intents}
  \figurespace{}
\end{figure}


\sys{} relocates parameters only when there is (at one point in time)
\emph{exactly one} node with active intent, and this node does currently not hold the
parameter. It does not relocate a parameter while multiple nodes have active intent,
even if the intent of the current owner expires. Figure
Figure~\ref{fig:relocate-while-intents} illustrates this approach. \sys{} employs this approach because
relocating in the presence of replicas would require (i) to update routing
information on each replica holder (see Section~\ref{sec:sys:routing}) and (ii)
to transfer intent information to another node (see
Section~\ref{sec:sys:decision-making-node}).


\begin{table*}
  \caption{ML tasks, models, and datasets.}
  \label{tab:datasets}
  \centering
  \begin{threeparttable}
  \begin{tabular}{llrrrlrr} \toprule
    Task & \multicolumn{4}{c}{Model parameters} & \multicolumn{3}{c}{Data}
    \\
    \cmidrule(lr){2-5} \cmidrule(lr){6-8}
         & Model                     & \mbox{Keys}           & \hspace{-1cm}\mbox{Values} & Size & Dataset & \mbox{\hspace{-2.6cm}Data points} & Size
    \\ \midrule

    Knowledge graph embeddings\hspace{-0.2cm} & \multicolumn{1}{l}{ComplEx, dim.~\num{500}} & \SI{4.8}{M} & \SI{4.8}{B} & \SI{35.9}{GB} & Wikidata5M & \SI{21}{M} & \hspace{-0.3cm}\SI{317}{MB}
    \\

    Word vectors & \multicolumn{1}{l}{Word2Vec, dim.~\num{1000}} & \SI{1.9}{M} & \SI{1.9}{B} & \SI{14.0}{GB} & 1b word benchmark & \SI{375}{M} & \SI{3}{GB}
    \\

    Matrix factorization            & \multicolumn{1}{l}{Latent Factors, rank~1000} & \SI{11.0}{M}& \SI{11.0}{B} & \SI{163.9}{GB} & \rectM{} matrix, zipf 1.1 & \SI{1000}{M} & \SI{31}{GB}
    \\

    Click-through-rate prediction\hspace{-0.2cm} & \multicolumn{1}{l}{Wide \& Deep, dim.~\num{128}} & \SI{67}{M} & \SI{8.7}{B} & \SI{32.5}{GB} & Criteo Kaggle & \SI{46}{M} & \SI{7}{GB}
    \\

    Graph neural networks\hspace{-0.2cm} & \multicolumn{1}{l}{GCN, dim.~\num{500}} & \SI{111}{M} & \SI{111.1}{B} & \SI{413.7}{GB} & ogbn-papers100M & \SI{111}{M} & \hspace{-0.3cm}\SI{56}{GB}
    \\

    \bottomrule
  \end{tabular}
  \end{threeparttable}
\end{table*}


\section{Task details}
\label{sec:task-details}

\textbf{Knowledge graph embeddings. } KGE models
learn algebraic representations of the entities and relations in a knowledge
graph. For example, these representations have been applied successfully to
infer missing links in knowledge graphs~\cite{kgcompletion}. This task, based
on Liu et al.~\cite{analogy}, trains ComplEx~\cite{complex} (one of the most popular KGE
models) embeddings using SGD with AdaGrad~\cite{adagrad} and negative
sampling~\cite{kgetraining, analogy}. To generate negative samples, both the
subject and the object entity of a positive triple are perturbed
$n_{\operatorname{neg}}$ times, by drawing random entities from a uniform
distribution over all entities (we used a common setting of
$n_{\operatorname{neg}}=100$~\cite{kgetraining}). We used the Wikidata5M
dataset~\cite{wikidata5m}, a real-world knowledge graph with \num{4818679}
entities and \num{828} relations, and a common embedding size of
\num{500}~\cite{kgetraining}. We partitioned the subject--relation--object
triples of the dataset to the nodes randomly, as done in~\cite{libkge-dist}. We
used LibKGE~\cite{libkge} (commit 3146885) to evaluate models and report the
\emph{mean reciprocal rank (filtered)} (MRRF) as model quality.

\textbf{Word vectors. } WVs are a language modeling technique in
natural language processing: each word of a vocabulary is mapped to a vector of
real numbers~\cite{word2vec,glove,elmo}. These vectors are useful as input for
many natural language processing tasks, for example, syntactic
parsing~\cite{syntactic-parsing} or question
answering~\cite{question-answering}. This task, based on Mikolov et al. \cite{word2vec}, uses
SGD with AdaGrad~\cite{adagrad} and negative sampling to train the skip-gram
Word2Vec~\cite{word2vec} model (dimension~1000) on the One Billion Word
Benchmark~\cite{1b-word} dataset (with stop words of the Gensim~\cite{gensim}
stop word list removed). We used common model parameters~\cite{word2vec} for window
size (5), minimum count (1), and frequent word subsampling (0.01). We measured
model accuracy using a common analogical reasoning task of \num{19544} semantic
and syntactic questions~\cite{word2vec}.

\textbf{Matrix factorization. } Low-rank MF is a common
tool for analyzing and modeling dyadic data, e.g., in collaborative filtering
for recommender systems~\cite{mf-recsys}. This task, based on Teflioudi et al. \cite{dsgdpp},
uses SGD with AdaGrad~\cite{adagrad} to factorize a synthetic, Zipf-1.1
distributed \rectM{} dataset with 1b revealed cells, modeled after the Netflix
Prize dataset.\footnote{See \url{https://netflixprize.com/}. We use a synthetic
  dataset because the largest openly available dataset that we are aware of is
  only \SI{7.6}{GB} large.} Data points were partitioned to nodes by row and to
workers within a node by column. Each worker visited its data points by column
(to create locality in column parameter accesses), with random order of columns
and of data points within a column. We report the \emph{root mean squared error}
(RMSE) on the test set as metric for model quality.

\begin{figure*}
  \panelfiguresetup{}
  \centering
  \includegraphics[page=1,width=1.0\textwidth]{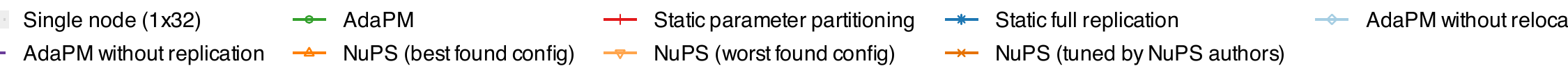}

  \begin{subfigure}[b]{0.33\textwidth}
    \centering
    \includegraphics[page=1,width=0.95\textwidth]{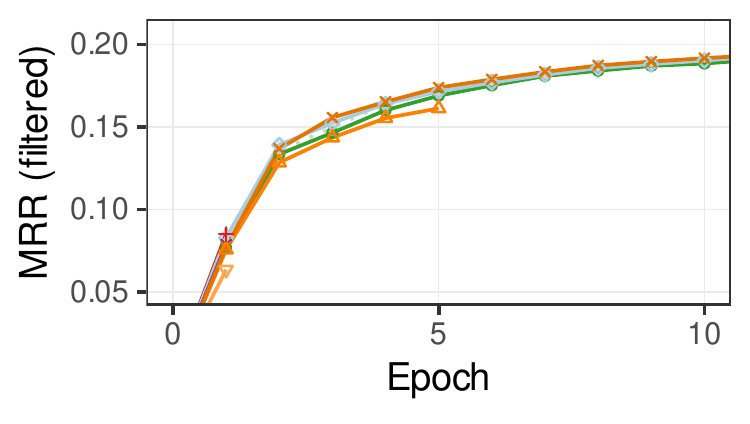}
    \caption{Knowledge graph embeddings (KGE)}
    \label{fig:ete:kge:epoch}
  \end{subfigure}%
  \begin{subfigure}[b]{0.33\textwidth}
    \centering
    \includegraphics[page=1,width=0.95\textwidth]{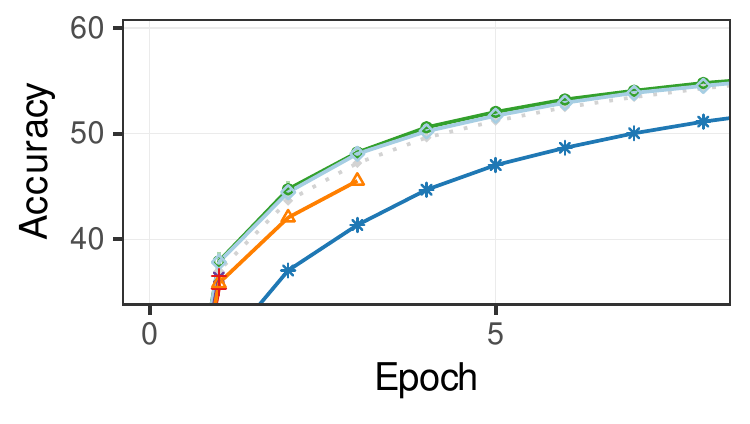}
    \caption{Word vectors (WV)}
    \label{fig:ete:wv:epoch}
  \end{subfigure}%
  \begin{subfigure}[b]{0.33\textwidth}
    \centering
    \includegraphics[page=1,width=0.95\textwidth]{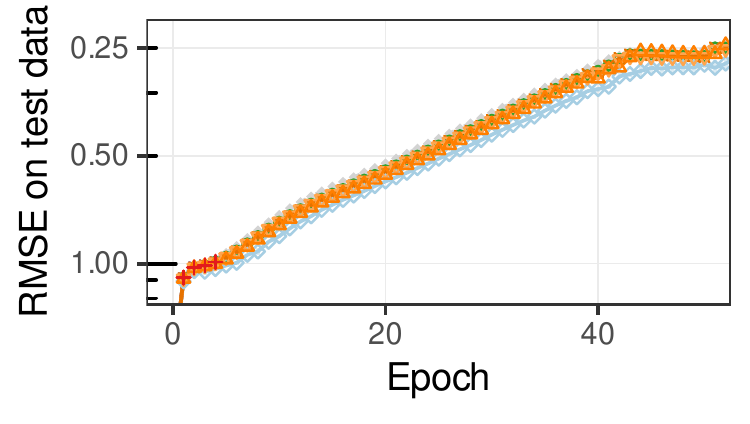}
    \caption{Matrix factorizaton (MF)}
    \label{fig:ete:mf:epoch}
  \end{subfigure}%

  \begin{subfigure}[b]{0.33\textwidth}
    \centering
    \includegraphics[page=1,width=0.95\textwidth]{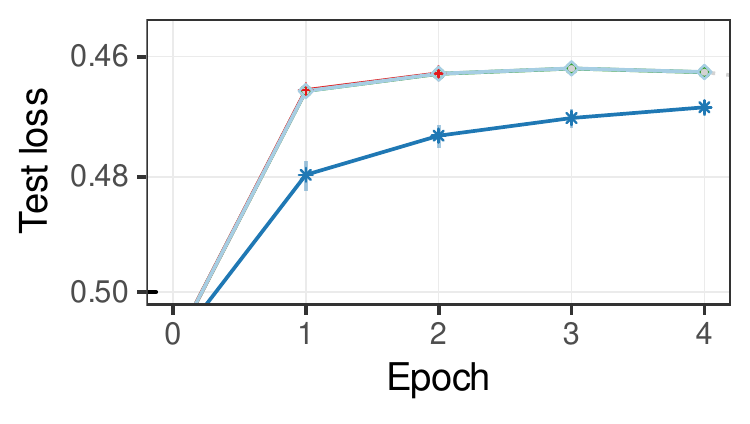}
    \caption{Click-through-rate prediction (CTR)}
    \label{fig:ete:ctr:epoch}
  \end{subfigure}%
  \begin{subfigure}[b]{0.33\textwidth}
    \centering
    \includegraphics[page=1,width=0.95\textwidth]{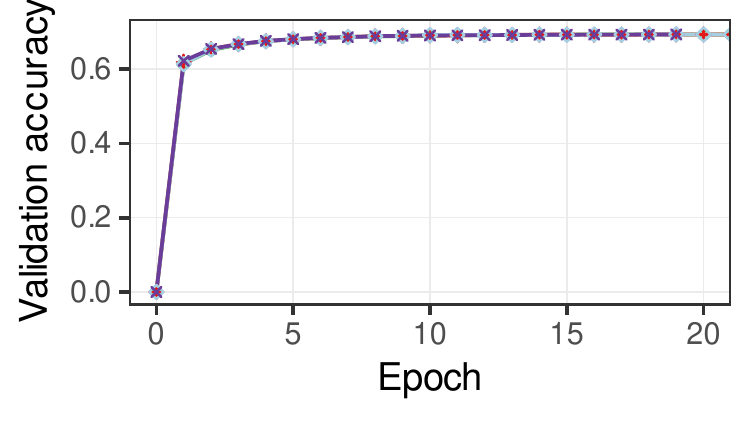}
    \caption{Graph neural networks (GNN)}
    \label{fig:ete:gnn:epoch}
  \end{subfigure}%

  \caption{Model quality \emph{per epoch}. }
  \label{fig:ete:epoch}
  \figurespace{}
\end{figure*}


\textbf{Graph neural networks. } GNNs are a class of artificial neural networks
that capture graph dependencies via message passing between the nodes of the
graph. This task, based on Wang et al.~\cite{dgl}, trains a 2-layer graph
convolutional network (GCN)~\cite{rgcn} for the node property prediction task
for the
ogbn-papers100M\footnote{\url{https://ogb.stanford.edu/docs/nodeprop/\#ogbn-papers100M}}
dataset from the open graph benchmark~\cite{open-graph-benchmark}, using
mini-batch SGD, neighbor sampling~\cite{graphsage}, and AdaGrad~\cite{adagrad}.
The task ignores the provided word embedding features and instead learns node
embeddings from scratch. We used an embedding dimension of \num{500}, a batch
size of 32 (per worker), and sample \num{16} neighbors per layer. We partitioned the graph to
the cluster nodes using METIS~\cite{metis} and report validation accuracy as
model quality.

\textbf{Click-through-rate prediction. } CTR models are common in online
advertising to evaluate ad performance. This task, based on Wang et al.~\cite{hugectr}, trains the
Wide \& Deep model~\cite{widedeep} (one of the most common CTR models) with
mini-batch SGD and AdaGrad~\cite{adagrad} on the Criteo Kaggle
dataset.\footnote{\url{https://www.kaggle.com/competitions/criteo-display-ad-challenge/data}}
We used an embedding dimension of \num{128}, a batch size of \num{128} (per worker), and a
learning rate of \num{0.0001}. As Wang et al.~\cite{deepcross}, we used the
first 6 days of data as the training data, and randomly split the 7th day into
validation and test data sets. We randomly partitioned the training data to the
nodes and report the test loss as model quality.

\begin{figure}
  \panelfiguresetup{}
  \centering
  \includegraphics[page=1,width=1.0\columnwidth]{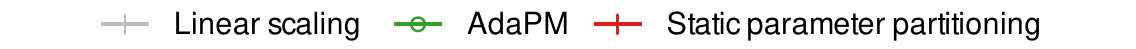}

  \begin{subfigure}[b]{.35\columnwidth}
    \centering
    \includegraphics[page=1,width=1.0\textwidth]{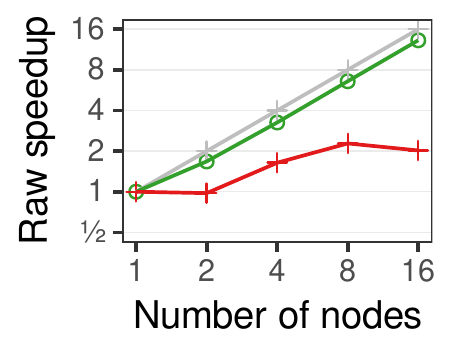}
    \caption{CTR: raw}
    \label{fig:scale:ctr}
  \end{subfigure}%
  \begin{subfigure}[b]{.324\columnwidth}
    \centering
    \includegraphics[page=1,width=1.0\textwidth]{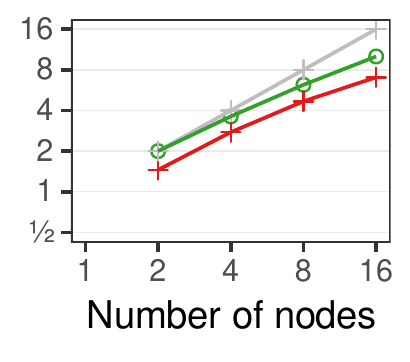}
    \caption{GNN: raw}
    \label{fig:scale:gnn}
  \end{subfigure}%

  \begin{subfigure}[b]{.35\columnwidth}
    \centering
    \includegraphics[page=1,width=1.0\textwidth]{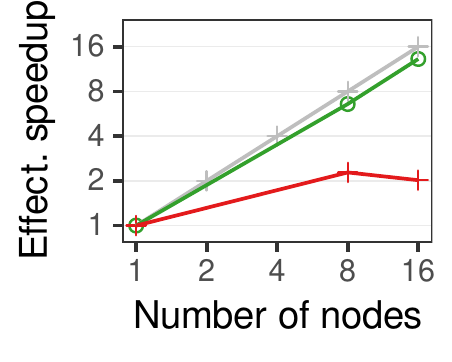}
    \caption{CTR: effective}
    \label{fig:eff-scale:ctr}
  \end{subfigure}%
  \begin{subfigure}[b]{.324\columnwidth}
    \centering
    \includegraphics[page=1,width=1.0\textwidth]{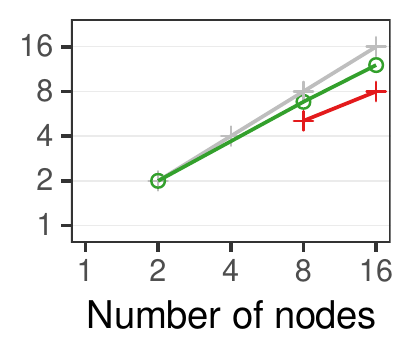} 
    \caption{GNN: effective}
    \label{fig:eff-scale:gnn}
  \end{subfigure}%

  \caption{Further scalability results (logarithmic axes). Raw (a-b) and effective speedup (c-d).}
  \label{fig:scale:appendix}
  \figurespace{}
\end{figure}


\textbf{Negative sampling.} For the tasks that use negative sampling (KGE and WV), we used
local sampling in NuPS and \sys{}. In static parameter partitioning, we used
independent sampling, as local sampling provides poor sampling quality when
combined with static parameter partitioning~\cite{nups}.

\textbf{Default intent signal offset.} In \sys{}, we used arbitrary large values
for the intent signal offset (\num{1000} data points in KGE, \num{2000}
sentences in WV, and \num{10000} data points in MF, \num{100} batches in DLRS
and GNN). When chosen large enough, the offset did not affect \sys{}'s performance
(see Section~\ref{sec:exp:action-timing}).

\begin{figure}
  \centering
  \panelfiguresetup{}
  \includegraphics[page=1,width=1.0\columnwidth]{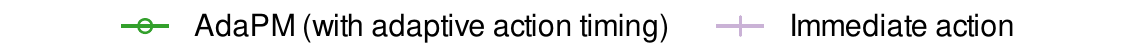}

  \begin{subfigure}[b]{0.362\columnwidth}
    \centering
    \includegraphics[page=1,width=1.00\textwidth]{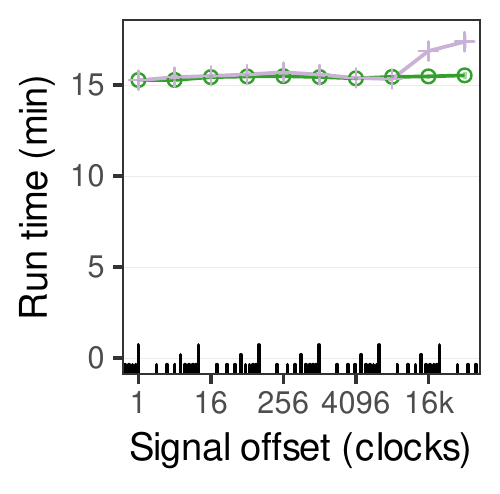}
    \caption{KGE: run time}
  \end{subfigure}%
  \begin{subfigure}[b]{0.3187\columnwidth}
    \centering
    \includegraphics[page=1,width=1.00\textwidth]{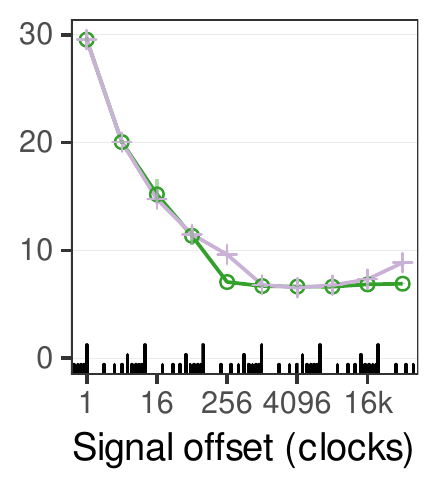}
    \caption{WV: run time}
  \end{subfigure}%
  \begin{subfigure}[b]{0.3187\columnwidth}
    \centering
    \includegraphics[page=1,width=1.00\textwidth]{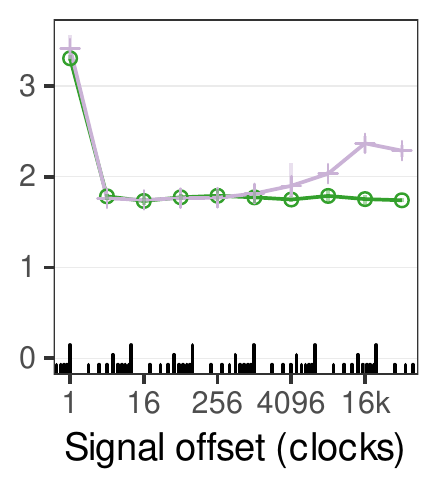}
    \caption{MF: run time}
  \end{subfigure}%

  \begin{subfigure}[b]{0.362\columnwidth}
    \centering
    \includegraphics[page=1,width=1.00\textwidth]{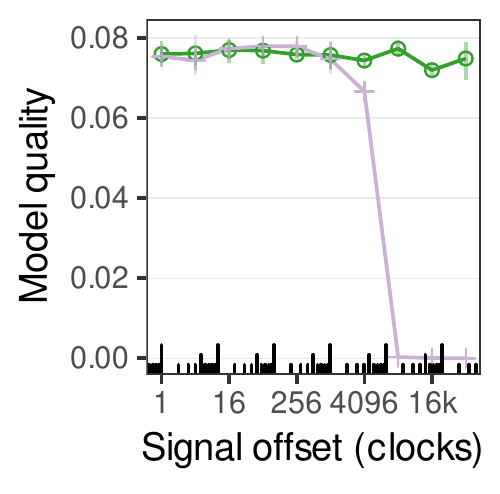}
    \caption{KGE: quality}
  \end{subfigure}%
  \begin{subfigure}[b]{0.3187\columnwidth}
    \centering
    \includegraphics[page=1,width=1.00\textwidth]{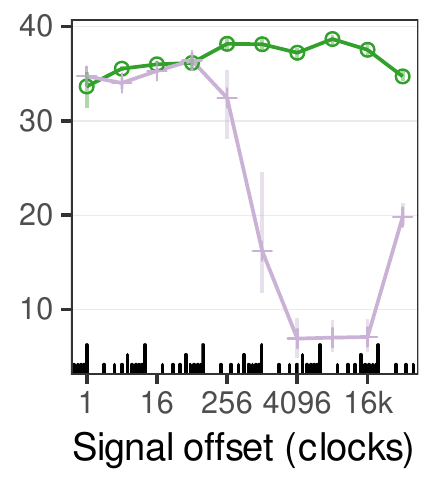}
    \caption{WV: quality}
  \end{subfigure}%
  \begin{subfigure}[b]{0.3187\columnwidth}
    \centering
    \includegraphics[page=1,width=1.00\textwidth]{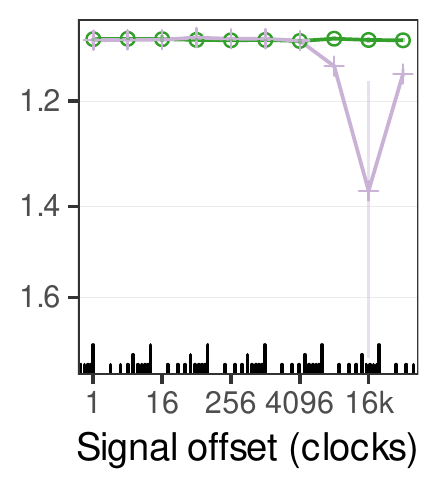}
    \caption{MF: quality}
  \end{subfigure}%
  
  \caption{The effect of adaptive action timing on epoch run time (a-c) and
    model quality after one epoch (d-f). }
  \label{fig:action-timing}
  \figurespace{}
\end{figure}


\section{NuPS Configurations}
\label{sec:nups-configurations}
The main performance hyperparameters of NuPS are (i) choosing the
management techniques for each parameter and (ii) specifying a relocation
offset.
We generated five configurations quasi-randomly using the Sobol sequence
implementation of Ax~\cite{ax}. For choosing techniques, we narrowed the search
range using the NuPS heuristic proposed by Renz-Wieland et al.~\cite{nups} (based on pre-computed
dataset frequency statistics) for each task. We generated one set of
configurations that replicate 0.01x--100x as many parameters (because
Renz-Wieland et al. \cite{nups} report that up to 64x deviations from the heuristic were beneficial). For the
relocation offset, we set the search space to 1--1000 as we found that offsets
of up to 512 can be beneficial (see Section~\ref{sec:exp:action-timing}). We use
the same set of configurations for the three tasks. We provide more details on
the search and the exact configurations online.


\newcommand{\rmark}[1]{
  \draw[draw=red,opacity=0.8,thick] (#1) rectangle +(0.09,0.24);
}

\newcommand{\trace}[4]{
  \begin{subfigure}[c]{1.0\columnwidth}
      \begin{tikzpicture}[>=stealth',auto]

        \node [anchor = south west] at (0,0) {\includegraphics[clip, trim=4cm 0.2cm 20cm 0.1cm, width=1.00\columnwidth]{traces/kge/#1.pdf}};

        #4

      \end{tikzpicture}

    \caption{#2}
    \label{fig:traces:#3}
  \end{subfigure}%
}

\begin{figure}

  \center
  \resizebox{0.5\columnwidth}{!}{
  \includegraphics[clip, trim=0.95cm 0.02cm 1.2cm 0cm, width=0.65\columnwidth]{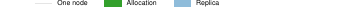}
  }

  \trace{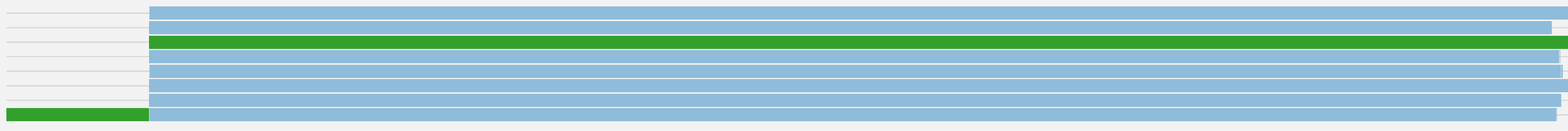}{Hot spot parameter (most frequent parameter)}{most-frequent}{}
  \trace{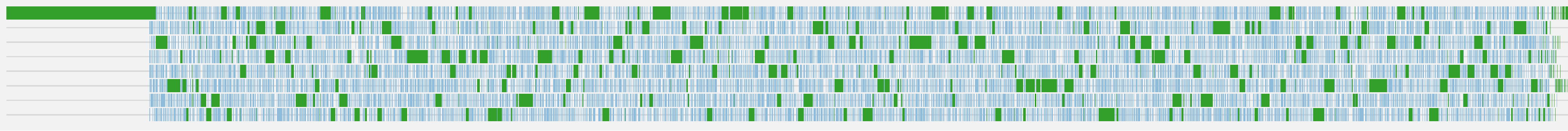}{Hot spot parameter (99.997\% frequency quantile)}{hotspot}{}
  \trace{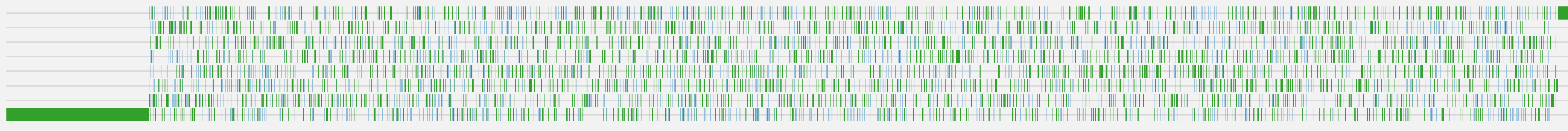}{Hot spot parameter (99.991\% frequency quantile)}{hotspottwo}{}
  \trace{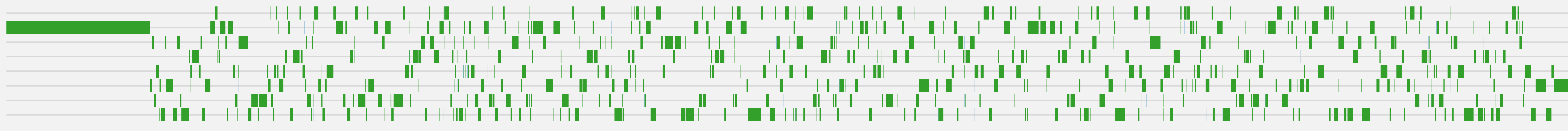}{Frequent parameter (99.95\% frequency quantile)}{frequent}{
    \rmark{1.18,0.46}
    \rmark{2.08,0.105}
  }
  \trace{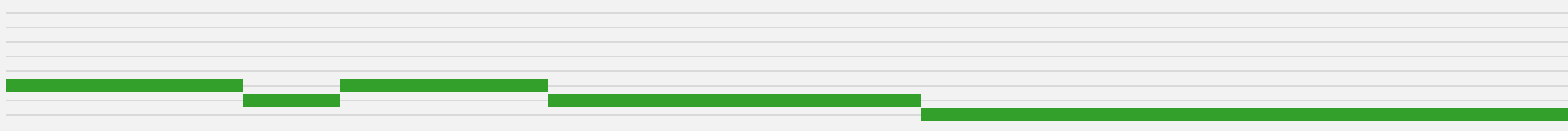}{Typical parameter (median frequency)}{median}{}

  \caption{Parameter management for selected parameters in the KGE task. Each
    row corresponds to one of 8 nodes. The red boxes indicate two short-lived
    replicas.}
\label{fig:traces}
\figurespace{}
\end{figure}


\section{\sys{} in Action (Figure~\ref{fig:traces})}
\label{sec:exp:in-action}

\sys{} decides dynamically and automatically between relocation and replication.
To explore how \sys{} actually manages parameters, we traced \sys{}'s parameter
management. Figure~\ref{fig:traces} depicts parameter management for selected
parameters during the first half of the first epoch of KGE training on 8 nodes.
\textbf{ \sys{} managed extreme hot spots and extreme cold spots the same way a
  multi-technique PS does, but used more efficient approaches for parameters
  between the extremes.}
  
\textbf{The extremes. } NuPS decides statically how to manage a parameter.
The NuPS heuristic would use static full replication for the first
three depicted parameters (Figures~\ref{fig:traces:most-frequent}, and
\ref{fig:traces:hotspot}, and \ref{fig:traces:hotspottwo}) and relocation for
the others. \sys{} ended up managing the two extremes---i.e., the extreme hot
spot (Figure~\ref{fig:traces:most-frequent}) and the rarely accessed parameter
(Figure~\ref{fig:traces:median})---as NuPS would.

\textbf{Between the extremes. } For the parameters between these two extremes,
\sys{} took more fine-grained approaches than NuPS. For example, for the parameter
in Figure~\ref{fig:traces:hotspot}, \sys{} maintained replicas exactly while
they were needed. Concretely, the gray areas in Figures~\ref{fig:traces:hotspot}
and~\ref{fig:traces:hotspottwo} indicate periods in which \sys{} communicates less
than NuPS: in contrast to \sys{}, NuPS would maintain replicas during these periods.
For the parameter in Figure~\ref{fig:traces:frequent}, \sys{} created
(short-lived) replicas whenever multiple nodes accessed this parameter
concurrently. These short-lived replicas are barely visible in the figure, so we
highlighted two of them with red boxes. The short-term replicas prevented
workers from having to access the parameter remotely. In contrast, NuPS would
manage this parameter exclusively by relocation, such that workers are slowed
down by remote parameter accesses.


\fi


\end{document}